\documentclass[mathfont=cm,accepted,hidelinks]{uai2021} 
\RequirePackage{scrlfile}
\usepackage{subfigure}
\pdfoutput=1

\usepackage{xr}
\usepackage{svg}
\usepackage{parcolumns}
\usepackage{setspace}
\usepackage{tabularx}
\usepackage{wrapfig}
\usepackage{cancel}
\usepackage{enumitem}
\usepackage{natbib}
\setcitestyle{authoryear,open={(},close={)}}

\usepackage[english]{babel}
\usepackage[utf8]{inputenc} 
\usepackage[T1]{fontenc}    
\usepackage{booktabs}
\usepackage{tabularx}
\usepackage{algorithm}
\usepackage{algorithmic,amssymb,xspace,nicefrac,etoolbox}
\usepackage{setspace}

\usepackage[normalem]{ulem}

\usepackage{times}
\usepackage{url}
\usepackage{harpoon}
\usepackage{amssymb}
\usepackage{amsmath}
\usepackage{graphicx}
\usepackage{sidecap}

\usepackage{courier}
\usepackage{color}
\usepackage{caption}
\usepackage{lscape}
\usepackage{tikz}
\usepackage{PTSansNarrow}

\usetikzlibrary{matrix}

\usepackage{algorithm}
\usepackage{algorithmic,amssymb,xspace,nicefrac,etoolbox}

\usepackage{pgfplots}

\usepackage[acronym,smallcaps,nowarn,section,nogroupskip,nonumberlist,shortcuts]{glossaries}
\usepackage[nameinlink,capitalise]{cleveref}

\usetikzlibrary{arrows,intersections, decorations.markings}

\usepackage{amsthm}
\usepackage{booktabs}
\usepackage{nicefrac}       
\usepackage{microtype}      
\usepackage{amsfonts}       
\usepackage{harpoon}
\usepackage{parcolumns}

\makeatletter
\newcommand*{\addFileDependency}[1]{
  \typeout{(#1)}
  \@addtofilelist{#1}
  \IfFileExists{#1}{}{\typeout{No file #1.}}
}
\makeatother

\newtheorem{lemma}{Lemma}

\newcommand{\hl}[1]{{\color{blue}{#1}}}

\DeclareMathOperator{\E}{\mathbb{E}}

\newcommand{\vx}{x}
\newcommand{\vz}{z}
\def\KL{\mathrm{KL}}

\newcommand{\base}{g}
\newcommand{\z}{z}
\newcommand{\tpi}{\tilde{\pi}}

\DeclareMathOperator*{\argmin}{argmin}

\newcommand{\suffexp}{\phi}
\newcommand{\suffq}{T}
\newcommand{\qqq}{{q}}

\newcommand{\bmu}{{\mathbf{\mu}}}
\newcommand{\bSig}{{\mathbf{\Sigma}}}

\glsdisablehyper{}
\newacronym{AIS}{ais}{Annealed Importance Sampling}
\newacronym{PT}{pt}{parallel tempering}
\newacronym{SMC}{smc}{Sequential Monte Carlo}
\newacronym{KM}{km}{Kolmogorov-Nagumo}
\newacronym{BQ}{bq}{Bayesian Quadrature}
\newacronym{AUC}{auc}{area under the curve}
\newacronym{BAR}{bar}{Bennett's Acceptance Ratio}
\newacronym{BDMC}{bdmc}{Bidirectional Monte Carlo}
\newacronym{JS}{js}{Jensen-Shannon}
\newacronym{CFT}{cft}{Crooks's Fluctuation Theorem}
\newacronym{ESS}{ess}{effective sample size}
\newacronym{ELBO}{elbo}{evidence lower bound}
\newacronym{EUBO}{eubo}{evidence upper bound}
\newacronym{HMC}{hmc}{Hamiltonian Monte Carlo}
\newacronym{IB}{ib}{Information Bottleneck}
\newacronym{IS}{is}{importance sampling}
\newacronym{IWAE}{iwae}{importance weighted autoencoder}
\newacronym{KL}{kl}{Kullback Leibler}
\newacronym{MCMC}{mcmc}{Markov Chain Monte Carlo}
\newacronym{RD}{r-d}{rate-distortion}
\newacronym{RWS}{rws}{reweighted wake-sleep}
\newacronym{SGD}{sgd}{stochastic gradient descent}
\newacronym{SNIS}{snis}{self-normalized importance sampling}
\newacronym{TI}{ti}{thermodynamic integration}
\newacronym{TVI}{tvi}{thermodynamic variational inference}
\newacronym{TVO}{tvo}{thermodynamic variational objective}
\newacronym{VAE}{vae}{variational autoencoder}
\newacronym{VI}{vi}{variational inference}
\newacronym{VIMCO}{vimco}{variational inference for Monte Carlo objectives}
\newacronym{WS}{ws}{wake-sleep}

\title{q-Paths: Generalizing the Geometric Annealing Path using Power Means}

\author[1,5]{
    Vaden Masrani$^{1*}$,
    Rob Brekelmans$^{2*}$,
    Thang Bui$^3$,\\
    {\bf
    Frank Nielsen$^4$,
    Aram Galstyan$^2$,
    Greg Ver Steeg$^2$,
    Frank Wood
    }}

\affil{%
    University of British Columbia,$^{2}$USC Information Sciences Institute
}
\affil[3]{%
    University of Sydney, $^{4}$Sony CSL,$^{5}$MILA, $^{*}$Equal Contribution
}
\affil[ ]{
    {\tt \{vadmas,fwood\}@cs.ubc.ca},
    {\tt \{brekelma,galstyan,gregv\}@isi.edu},
    {\tt thang.bui@sydney.edu.au},
    {\tt frank.nielsen@acm.org}
}

\begin{document}

\maketitle

\begin{abstract}

    Many common machine learning methods involve the geometric annealing path, a sequence of intermediate densities between two distributions of interest constructed using the geometric average.   While alternatives such as the moment-averaging path have demonstrated performance gains in some settings, their practical applicability remains limited by exponential family endpoint assumptions and a lack of closed form energy function.   In this work, we introduce $q$-paths, a family of paths which is derived from a generalized notion of the mean, includes the geometric and arithmetic mixtures as special cases, and admits a simple closed form involving the deformed logarithm function from nonextensive thermodynamics.
    Following previous analysis of the geometric path, we interpret our $q$-paths as corresponding to a $q$-exponential family of distributions, and provide a variational representation of intermediate densities as minimizing a mixture of $\alpha$-divergences to the endpoints.  We show that small deviations away from the geometric path yield empirical gains for Bayesian inference using Sequential Monte Carlo and generative model evaluation using Annealed Importance Sampling. 

\end{abstract}

\newcolumntype{L}{>{\centering\arraybackslash}m{2.5cm}}

\begin{figure*}[ht]
    \centering
    \scalebox{0.92}{
        \begin{tabular}{Lcc}
        \toprule
        &  \textbf{Geometric Path}                     & \textbf{$q$-Path}  \\ \midrule 
        \textbf{Closed Form}     & $\tilde{\pi}_{\beta}(\vz) =  \pi_0(\vz)^{1-\beta} \tpi_1(\vz)^\beta $ & $\tpi_{\beta,q}(\vz) =  \big[ (1-\beta) \,  \pi_0(\vz)^{1-q} + \beta \, \tpi_1(\vz)^{1-q} \big]^{\frac{1}{1-q}} $ \\[12pt]
        \textbf{Log Linear}     & $ \tilde{\pi}_{\beta}(\vz) = \exp\{ (1-\beta)\log\pi_0(\vz) + \beta\log\tpi_1(\vz) \} $ & $\tpi_{\beta,q}(\vz) = \exp_q \{ (1-\beta) \, \ln_q \pi_0 (\vz)  + \beta \, \ln_q \tpi_1 (\vz) \}$ \\[12pt]
        \textbf{Exponential Family}  & $\tilde{\pi}_{\beta}(\vz) = \pi_0(\vz) \exp \left\{ \, \beta  \cdot  \log\frac{\tilde{\pi}_1(\vz)}{\pi_0(\vz)}\right\}$                               & $\tpi_{\beta,q}(\vz) = \pi_0(\vz)  \exp_q \left\{ \beta \cdot \ln_q\frac{\tpi_1(\vz)}{\pi_0(\vz)}\right\}$ \\[16pt]
        \textbf{Variational Representation}     & $\pi_{\beta}(\vz) = \argmin\limits_{r}(1-\beta)D_{\KL}[r\|\pi_{0}] +\beta D_{\KL}[r\|\pi_{1}]$ & $\tpi_{\beta, q}(\vz)=\argmin\limits_{\tilde{r}}(1- \beta) D_{\alpha}[\tpi_{0}||\tilde{r}] +\beta D_{\alpha}[\tpi_{1}||\tilde{r}]$\\ \bottomrule
        \end{tabular}}
    \caption{Summary of $q$-paths (right) in relation to the geometric path (left). $q$-paths recover the geometric path as $q \to 1$ and $\alpha = 2q - 1$ in Amari's $\alpha$-divergence $D_{\alpha}$. The deformed logarithm $\ln_q$ and its inverse $\exp_q$ are defined in \cref{sec:q_definition}. } \label{fig:M1}
\end{figure*}

\section{Introduction}
Given a tractable and often normalized base distribution $\pi_0(\vz)$ and unnormalized target $\tpi_1(\vz)$, many statistical methods require a path $\gamma: [0, 1] \to \mathcal{P}$, where $\mathcal{P}$ is a family of unnormalized density functions with $\gamma(0) = \pi_0(\vz)$ and $\gamma(1) = \tpi_1(\vz)$.
For example, marginal likelihood estimation methods such as \gls{TI} \citep{ogata1989monte} or \gls{AIS} \citep{neal2001annealed}
and \gls{MCMC} methods such as parallel tempering \citep{earl2005parallel} and \gls{SMC} \citep{del2006sequential} typically use the geometric path with mixing parameter $\beta$,
\begin{align}
\tilde{\pi}_{\beta}(\vz) = \exp\left\{(1-\beta)\log \pi_0(\vz) + \beta\log \tpi_1(\vz)\right\},\label{eq:geo_path}
\end{align}
 In the Bayesian context, $\pi_0(\vz)$ and $\pi_1(\vz)$ can represent the prior and posterior distribution, respectively, in which case the geometric path amounts to tempering the likelihood term \citep{friel2008marginal, nguyenEfficientSequentialMonteCarlo2015a}. 

Previous work has demonstrated theoretical or empirical improvements upon the geometric path can be achieved, but the applicability of these methods remains limited in practice due to restrictive assumptions on the parametric form of the endpoint distributions. \citet{gelman1998simulating} derive an optimal path in distribution space but this is intractable to implement beyond toy examples.  The moment-averaging path of \citet{grosse2013annealing} demonstrates performance gains for partition function estimation in Restricted Boltzmann Machines, but is only applicable for endpoint distributions which come from an exponential family.  \citet{thang} proposed a path based on $\alpha$-divergence minimization  using an iterative projection scheme from \cite{minka2005divergence} which is also reliant on exponential family assumptions.

In this work, we propose $q$-paths, which can be constructed between arbitrary endpoint distributions and admit a closed form that can be used directly for \gls{MCMC} sampling
\begin{align}
    \tpi_{\beta,q}(\vz) &= \bigg[ (1-\beta) \,  \pi_0(\vz)^{1-q} + \beta \, \tpi_1(\vz)^{1-q} \bigg]^{\frac{1}{1-q}}\label{eq:qpath_mix_form}
\end{align}
Our $q$-paths adapt the $\alpha$-integration of \citet{amari2007integration} to the problem of annealing between two unnormalized densities, with our notation $q$ intended to highlight connections with the deformed logarithm and exponential functions from nonextensive thermodynamics \citep{tsallis2009introduction, naudts2011generalised}.    $q$-paths may be viewed as taking the generalized mean \citep{kolmogorov1930, de2016mean} of the endpoint densities according to a mixing parameter $\beta$ and monotonic transformation function $\ln_q(u) = \frac{1}{1-q}( u^{1-q} - 1)$.   As $q \rightarrow 1$, we recover the natural logarithm and geometric mean in \cref{eq:geo_path}, while the arithmetic mean corresponds to $q=0$.

As previous analysis of the geometric path revolves around the exponential family of distributions \citep{grosse2013annealing, brekelmans2020tvo, brekelmans2020lref}, we show in Sec. \ref{sec:path_exp_fam} that our proposed paths have an interpretation as a $q$-exponential family of density functions
\begin{align}
    \tpi_{\beta,q}(\vz) = \pi_0(\z) \, \exp_q \left\{ \beta \cdot  \ln_q \frac{\tpi_1(\vz)}{ \pi_0(\vz)} \right\}. \label{eq:qpath_exp_form}
    \end{align}
\citet{grosse2013annealing} show that intermediate distributions along the geometric and moment-averaged paths correspond to the solution of a weighted forward or reverse \textsc{kl} divergence minimization objective, respectively.  In Sec. \ref{sec:vrep_breg}, we generalize these variational representations to $q$-paths, showing that $\tpi_{\beta,q}(\vz)$ minimizes the expected $\alpha$-divergence to the endpoints for an appropriate mapping between ${q \text{ and } \alpha}$.

Finally, we highlight several implementation considerations in Sec. \ref{sec:experiments}, observing that $q = 1-\delta$ for small $\delta$ appears most useful both for qualitative mixing behavior and numerical stability.    We provide a simple heuristic for setting an appropriate value of $q$, and find that $q$-paths can yield empirical gains for Bayesian inference using \gls{SMC} and marginal likelihood estimation for generative models using \gls{AIS}.

\begin{figure*}[ht]
    \centering
    \subfigure[$q=0$]{\includegraphics[trim={0 1.2cm 0 1cm},clip,width=0.24\textwidth]{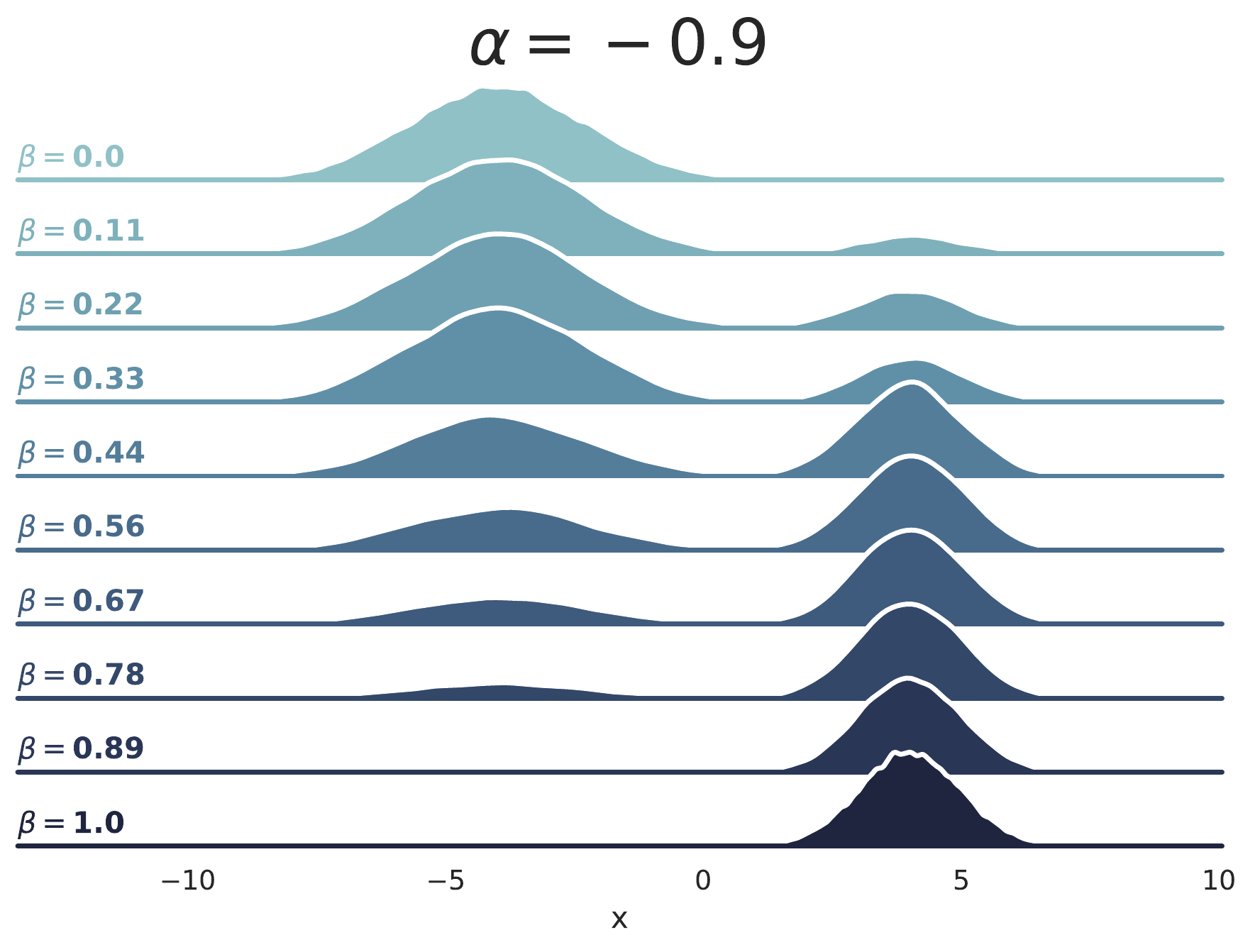}}
    \subfigure[$q=0.5$]{\includegraphics[trim={0 1.2cm 0 1cm},clip,width=0.24\textwidth]{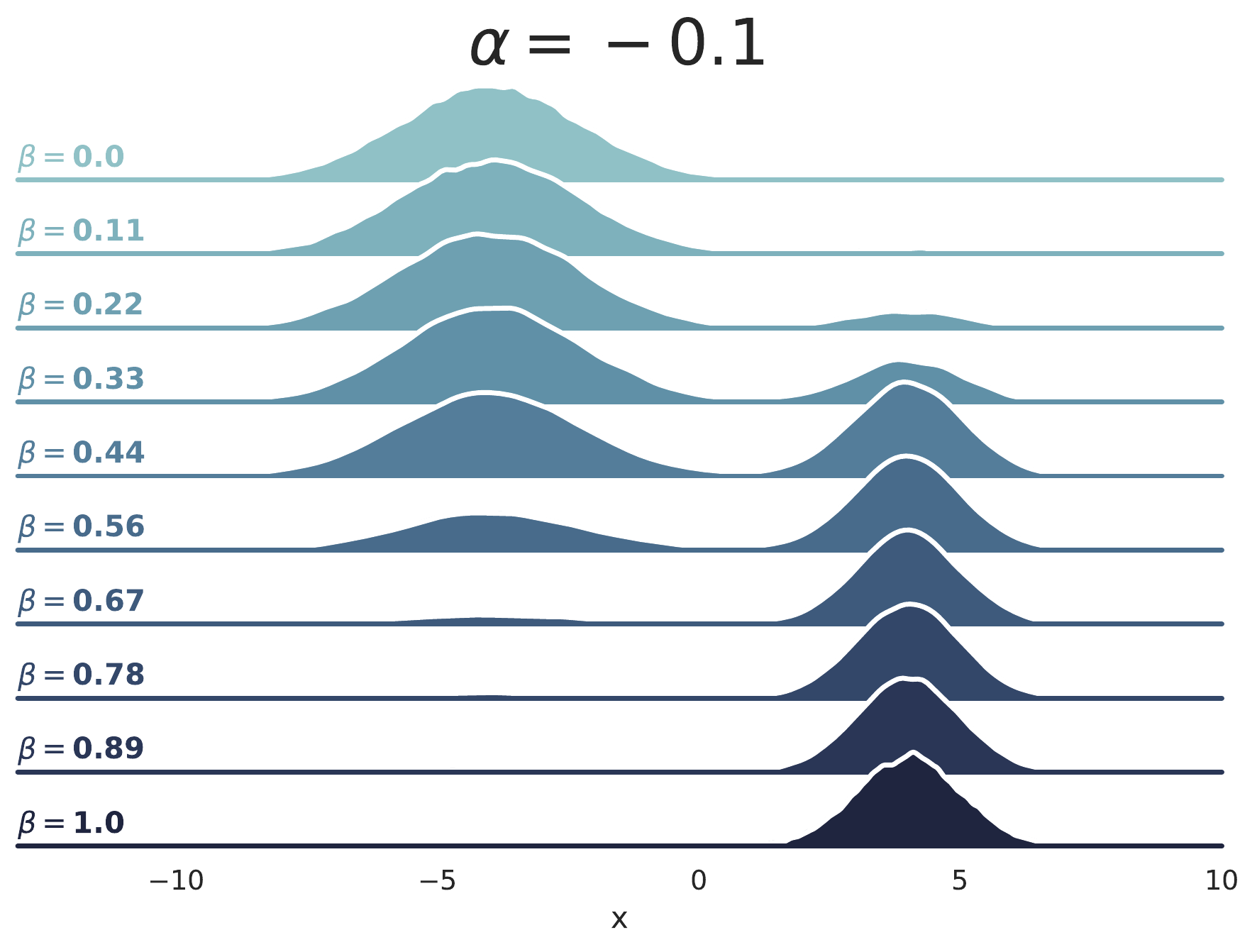}}
    \subfigure[$q=0.9$]{\includegraphics[trim={0 1.2cm 0 1cm},clip,width=0.24\textwidth]{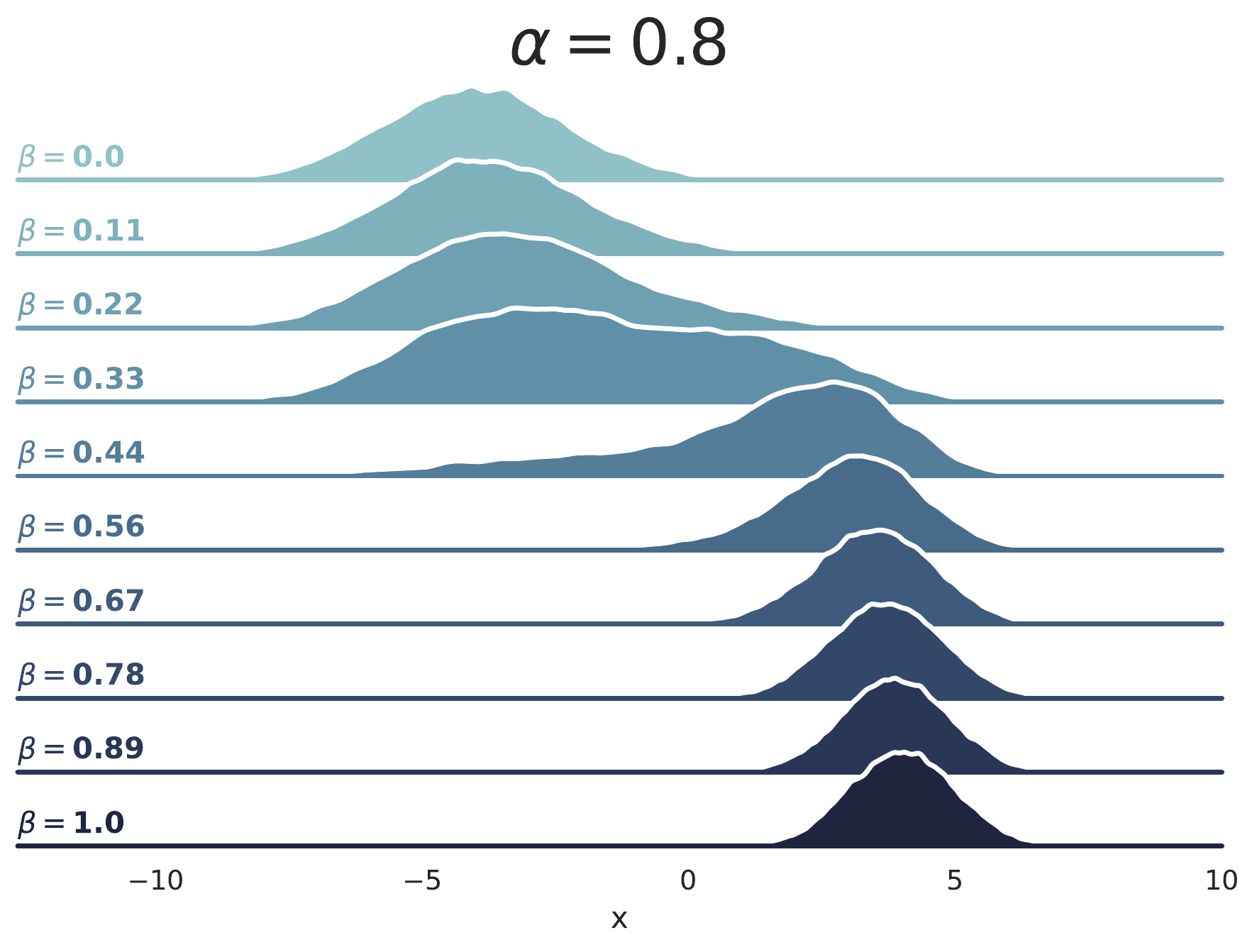}}
    \subfigure[$q=1$]{\includegraphics[trim={0 1.2cm 0 1cm},clip,width=0.24\textwidth]{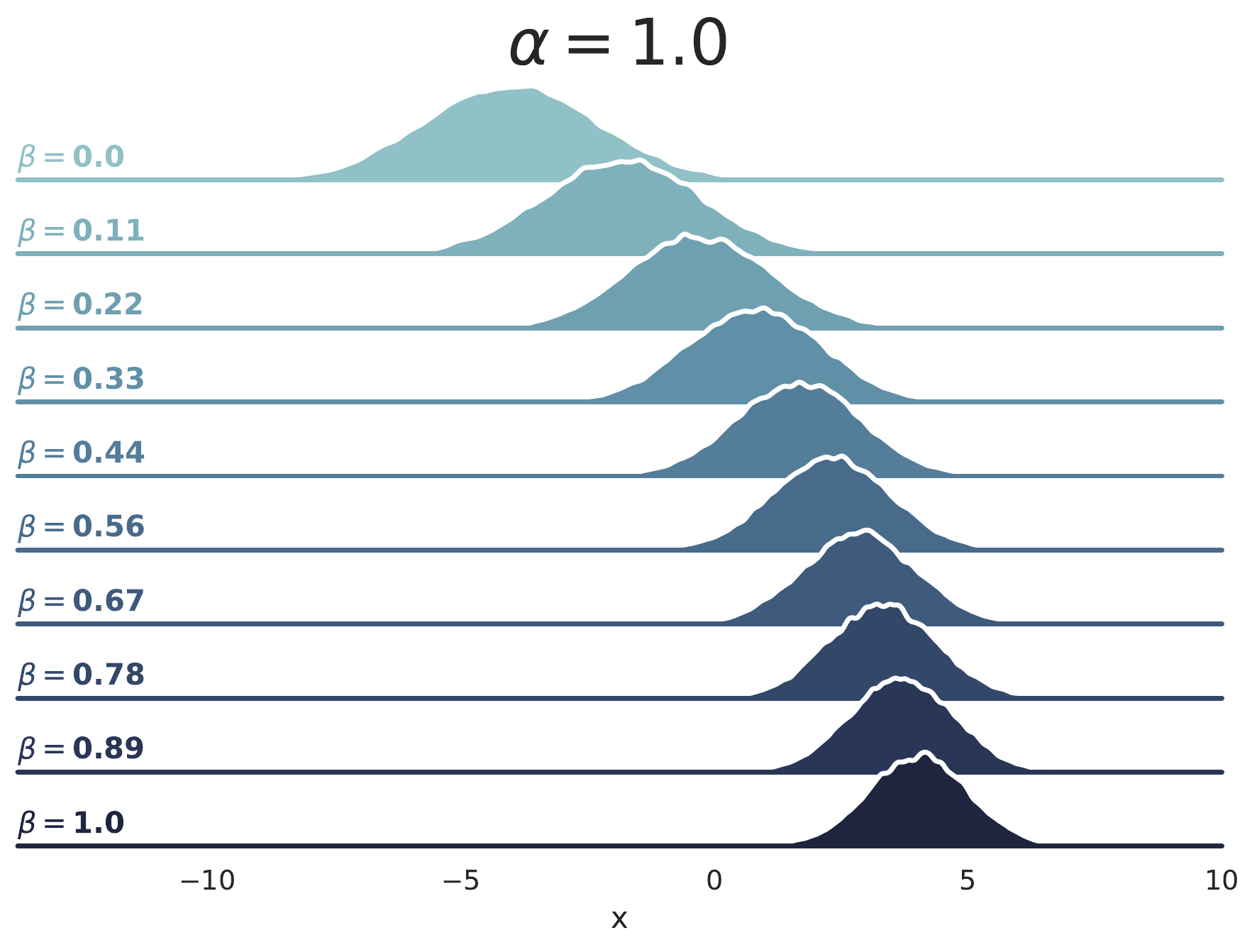}}
    \vspace*{-.2cm}
    \caption{Intermediate $q$-path densities between $\mathcal{N}(-4, 3)$ and $\mathcal{N}(4,1)$, with 10 equally spaced $\beta$.    For low $q$, the $q$-path approaches a mixture distribution at $q=0$, and becomes the geometric mixture parameterized by $\beta$ at $q=1$.}
    \label{fig:q_path}
\end{figure*}
\section{Background}

\subsection{Geometric annealing Path}
\label{sec:geo_path}
The geometric mixture path is the most ubiquitous method for specifying a set of intermediate distributions between a tractable base distribution $\pi_0$ and unnormalized target $\tpi_1$,
\begin{align}
    \pi_{\beta} (\vz) &= \frac{\pi_0(\vz)^{1-\beta} \, \tilde{\pi}_1(\vz)^{\beta}}{Z(\beta)}, \quad \text{where} \label{eq:geopath_} \\
    Z(\beta) &= \int \pi_0(\vz)^{1-\beta} \, \tilde{\pi}_1(\vz)^{\beta} d\vz.\label{eq:geopath0}
\end{align}

The geometric path may also be written as an exponential family of distributions, with natural parameter $\beta$ and sufficient statistic $\suffq(\vz) = \log \tpi_1(\vz) / \pi_0(\vz)$ corresponding to the log importance ratio.  
We follow \citet{grunwald2007minimum, brekelmans2020tvo, brekelmans2020lref} in referring to this as a \textit{likelihood ratio exponential family},  with
\begin{align} 
    \pi_{\beta}(\vz)&= \pi_0(\vz) \exp \bigg\{ \, \beta  \cdot  \log \frac{\tilde{\pi}_1(\vz)}{\pi_0(\vz)} \,   - \psi( \beta) \bigg\}\label{eq:lkd_ratio_fam}\\
    \psi(\beta) &:= \log Z(\beta) =  \log \int \pi_0(\vz)^{1-\beta} \tilde{\pi}_1(\vz)^{\beta} d\vz. \label{eq:lkd_ratio_partition}
\end{align}
It is often more convenient to work with \cref{eq:lkd_ratio_fam}, because one gains access to known exponential family properties that are not apparent from \cref{eq:geopath_} \citep{grosse2013annealing, brekelmans2020tvo,brekelmans2020lref}. 
In \cref{sec:path_exp_fam} we provide an analogous interpretation for $q$-paths in terms of $q$-exponential families.

\subsection{Moment Averaging Path}
Previous work \citep{grosse2013annealing} considers alternative annealing paths in the restricted setting where $\pi_0(\vz)$ and $\pi_1(\vz)$ are members of the same exponential family, with parameters $\theta_0$ and $\theta_1$ respectively.  Writing the base measure as $g(\vz)$ and sufficient statistics as $\phi(\vz)$,
\begin{align}
    \pi_{\theta}(\vz) = g(\vz) \exp \{ \theta \cdot \phi(\vz) - \psi(\theta) \} \label{eq:std_exp_fam}
\end{align}

\citet{grosse2013annealing} propose the \textit{moment-averaged} path based on the dual or `moment' parameters of the exponential family, which correspond to the expected sufficient statistics
\begin{align}
    \eta(\theta) = \frac{d \psi(\theta)}{d \theta}  =  \langle \mathbb{E}_{\pi_{\theta}}\left[ \phi_j(\vz) \right] \rangle_{j=1}^N \, , \label{eq:dpsi_dtheta}
\end{align}
with $\langle \cdot \rangle$ indicating vector notation and $\psi(\theta)$ denoting the log partition function of \cref{eq:std_exp_fam}.
In minimal exponential families, the sufficient statistic function $\eta(\theta)$ is a bijective mapping
between a natural parameter vector and dual parameter vector \citep{wainwrightjordan}.

The moment-averaged path is defined using a convex combination of the dual parameter vectors \citep{grosse2013annealing} 
\begin{align}
\eta(\theta_{\beta}) = (1-\beta) \, \eta(\theta_0) + \beta \, \eta(\theta_1) \, . \label{eq:moments_path}
\end{align}
To solve for the corresponding natural parameters, we calculate the Legendre transform, or a function inversion $\eta^{-1}$.
\begin{align}
    \theta_{\beta} = \eta^{-1}\big((1-\beta) \, \eta(\theta_0) + \beta \, \eta(\theta_1)\big) \label{eq:moments_path_theta} \,.
\end{align}
This inverse mapping is often not available in closed form and can itself be a difficult estimation problem \citep{wainwrightjordan, grosse2013annealing}, which limits the applicability of the moment-averaged path in practice.

\subsection{q-Deformed Logarithm / Exponential}\label{sec:q_definition}
\label{sec:qdeformed}
While the standard exponential arises in statistical mechanics via the Boltzmann-Gibbs distribution, \citet{tsallis1988possible} proposed a generalized exponential which has formed the basis of nonextensive thermodynamics and found wide application in the study of complex systems \citep{gell2004nonextensive, tsallis2009introduction}.

Consider modifying the integral representation of the natural logarithm $\ln u := \int_1^u \frac{1}{x}dx$ using an arbitrary power function
\begin{align}
    \ln_q u = \int_1^u \frac{1}{x^q}dx.\label{eq:lnq_int}
\end{align}
Solving \cref{eq:lnq_int} yields the definition of the $q$-logarithm
\begin{align}
    \ln_q(u) := \frac{1}{1-q} \left( u^{1-q} - 1 \right)  \label{eq:lnq} \, .
\end{align}
We define the $q$-exponential as the inverse of $q$-logarithm $\exp_q(u) := \ln_q^{-1}(u) $
\begin{align}
    \exp_q(u) = \big[ 1 + (1-q) \, u \big]_{+}^{\frac{1}{1-q}} \, , \label{eq:expq}
\end{align}
where $[x]_{+}= \max\{0, x\}= \textsc{relu}(x)$ ensures that $\exp_q(u)$ is non-negative and fractional powers can be taken for $q<1$, and thus restricts the domain where $\exp_q(u)$ takes nonzero values to $u > -1/(1-q)$.  We omit this notation in subsequent derivations because our $q$-paths in \cref{eq:qpath_mix_form} take non-negative densities as arguments for the $1/(1-q)$ power.

Note also that both the $q$-log and $q$-exponential recover the standard logarithm and exponential function in the limit,
\begin{align*}
&\lim_{q \rightarrow 1} \ln_q(u) & &&&& & \lim_{q \rightarrow 1} \exp_q(u)\\
=&\lim_{q \rightarrow 1} \frac{\frac{d}{dq} (u^{1-q} -1)}{\frac{d}{dq} (1-q)}  & &&&& =&\lim_{q \rightarrow 1}\left[1 + (1 - q) \cdot u \right]^{\frac{1}{1-q}}\\
=& \frac{ - \log u  \cdot u^{1-q}}{-1} \bigg|_{q=1} & &&&& =&\lim_{n \rightarrow \infty}\left[1 + \frac{u}{n}\right]^{n}\\
=&\log (u) & &&&& :=&\exp(u).
\end{align*}
In \cref{sec:path_exp_fam} we use this property to show $q$-paths recover the geometric path as $q \to 1$.

\begin{figure*}[ht]
    \centering
    \subfigure{\includegraphics[trim={0 0cm 0 0cm},clip,width=\textwidth]{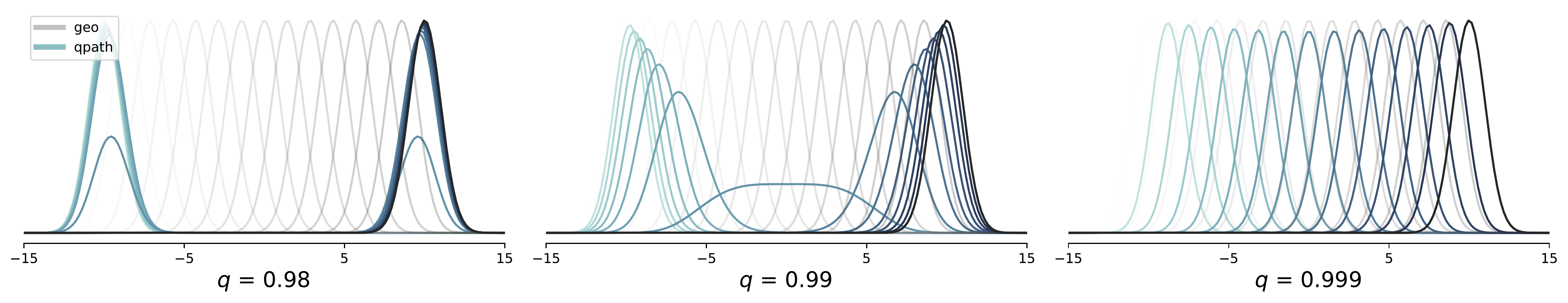}}
    \vspace*{-.35cm}
    \caption{$q$-paths between $\mathcal{N}(-10, 1)$ and $\mathcal{N}(10,1)$, which are notably more separated than those in Fig. \ref{fig:q_path}.  For difficult annealing problems such as those in our experiments, small deviations from the geometric path (grey) can achieve mass-covering behaviour (center), which is lost if the $q$-path too much resembles the arithmetic (left) or geometric mean (right).} 
    \label{fig:q_path_separated}
\end{figure*}

\section{q-Paths from Power Means} \label{sec:q_paths}
$q$-paths are derived using a generalized notion of the mean due to \citet{kolmogorov1930}. For any monotonic function $h(u)$, we define the \textit{generalized mean}
\begin{align}
\mu_{h}({\bf{u, w}} ) = h^{-1} \left(\sum_{i=1}^N w_i \cdot h(u_i)\right), \label{eq:abstract_mean}
\end{align}
where $\mu_{h}$ outputs a scalar given a normalized measure ${{\bf{w}} = (w_1, ..., w_N)}$ (with $\sum_{i=1}^N w_i = 1$) over a set of input elements ${\bf{u}} = (u_1, ..., u_N)$ \citep{de2016mean}.\footnote{The generalized mean is also referred to as the \textit{abstract}, \textit{quasi-arithmetic}, or \textit{Kolmogorov-Nagumo} mean in the literature.}

The generalized mean can be thought of as first applying a nonlinear transformation function to each input, applying the desired weights in the transformed space, and finally mapping back to the distribution space.

The geometric and arithmetic means are \textit{homogeneous}, that is, they have the linear scale-free property ${\mu_{h}(c \cdot {\bf{u}}, {\bf{w}}) = c \cdot \mu_{h}({\bf{u, w}})}$. \citet{hardy1953} shows the unique class of functions $h(u)$ that yield means with the homogeneity property are of the form
\begin{align}
h_{q}(u) =
\begin{cases}
    a \cdot u^{1-q} + b & q \neq 1 \\
    \log u \hfill 	&  q = 1
\end{cases}. \label{eq:alpha_abstract}
\end{align}
for any $a$ and $b$.  Setting ${a = b = 1 / (1-q)}$, we can recognize $h_q(u)$ as the deformed logarithm $\ln_q(u)$ from \cref{eq:lnq}.

Generalized means which use the class of functions $h_q(u)$ we refer to as \textit{power means}, and show in App. \ref{app:any_h} that for any choice of $a$ and $b$,
\begin{align}
 \mu_{h_q}({\bf{u, w}}) = \left[\sum_{i=1}^N w_i \cdot u_i^{1-q} \right]^{\frac{1}{1-q}}. \label{eq:powermean}
\end{align}
Notable examples include the arithmetic mean at $q = 0$, geometric mean as $q \rightarrow 1$, and the $\min$ or $\max$ operation as $q \rightarrow \pm \infty$.   For $q = \frac{1+\alpha}{2}$, $a = \frac{1}{1-q}$, and $b =0$, the function $h_{q}(u)$ matches the $\alpha$-representation in information geometry \citep{amari2016information}, and the resulting power mean over normalized probability distributions as input $\textbf{u}$ is known as the $\alpha$-integration \citep{amari2007integration}.

For annealing between unnormalized density functions, we propose the $q$-path of intermediate $\tilde{\pi}_{\beta,q}(\vz)$ based on the power mean.  Observing that the geometric mixture path in \cref{eq:geo_path} takes the form of a generalized mean for $h(u) = \ln(u)$, we choose the deformed logarithm
\begin{align}
    h_q(u) := \ln_q(u) \quad \quad h_q^{-1}(u) = \exp_q(u),
\end{align}
as the transformation function for the power mean. 
This choice will facilitate our parallel discussion of geometric and $q$-paths in terms of generalized logarithms and exponentials in \cref{sec:path_exp_fam}.

Using ${{\bf{u}} = (\pi_0, \tpi_1)}$ as the input elements and ${{\bf{w}} = (1-\beta, \beta)}$ as the mixing weights in \cref{eq:powermean}, we obtain a simple, closed form expression for the $q$-path intermediate densities
\begin{align}
    \tpi_{\beta,q} (\vz)
        &= \bigg[(1-\beta) \,  \pi_0 (\vz)^{1-q}  + \beta \, \tpi_1 (\vz)^{1-q} \bigg]^{\frac{1}{1-q}} \label{eq:qpath_mix_form22}
    \end{align}
Crucially, \cref{eq:qpath_mix_form22} can be directly used as an energy function in \gls{MCMC} sampling methods such as \gls{HMC} \citep{neal2011mcmc}, and our $q$-paths do not require additional assumptions on the endpoint distributions. 

Finally, to compare against the geometic path, we write the $q$-path in terms of the generalized mean in \cref{eq:abstract_mean}
\begin{align}
    \tilde{\pi}_{\beta,q} &= \exp_q\bigg\{(1-\beta) \, \ln_q \pi_0 (\vz)  + \beta \, \ln_q \tpi_1 (\vz)\bigg\} \, , \label{eq:lnq_mixture}
\end{align}
from which we can see that $\tilde{\pi}_{\beta,q}$ recovers the geometric path in \cref{eq:geo_path} as $q \to 1$, $\ln_q(u) \rightarrow \log(u)$, and $\exp_q(u) \rightarrow \exp(u)$.   Taking the deformed logarithm of both sides also yields an interpretation of the geometric or $q$-paths as $\ln$ or $\ln_q$-mixtures of density functions, respectively.

\begin{figure*}[ht]
    \centering
    \subfigure{\includegraphics[width=0.95\textwidth]{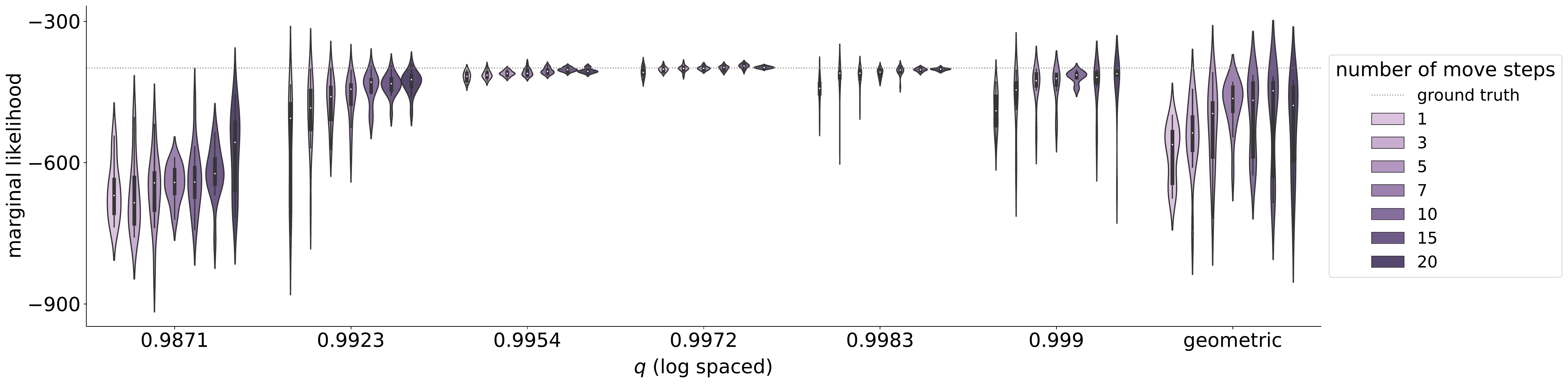}}
    \vspace*{-.2cm}
    \caption{SMC tempering using $q$-Paths on a binary regression model over 10 runs (cf. \cref{sec:experiment_details})). $q=0.9972$ outperforms the geometric path both in terms of marginal likelihood estimation and reduced variability across runs.}
    \label{fig:smc_sampling}
\end{figure*}

\section{q-Likelihood Ratio Exponential Families}\label{sec:path_exp_fam}
Similarly to \cref{eq:lkd_ratio_fam}, we relate $\tilde{\pi}_{\beta, q}$ to a $q$-exponential family with a single sufficient statistic and natural parameter $\beta$
\begin{align}
    \tilde{\pi}_{\beta, q}(\vz) &= \bigg[(1 - \beta) \pi_0(\vz)^{1 - q} + \beta \tpi_1(\vz)^{1 - q}\bigg]^\frac{1}{1 - q}\\
    &= \bigg[\pi_0(\vz)^{1 - q} + \hl{\beta}\big(\tpi_1(\vz)^{1 - q} - \hl{\pi_0(\vz)^{1 - q}}\big)\bigg]^\frac{1}{1 - q}\\
    &= \hl{\pi_0(\vz)} \left[\hl{1} + \beta\left(\left(\frac{\tpi_1(\vz)}{\hl{\pi_0(\vz)}}\right)^{1 - q} - \hl{1}\right)\right]^\frac{1}{1 - q}\\
    &= \pi_0(\vz) \left[1 + \hl{(1-q)} \, \beta \, \hl{\ln_q}\left( \frac{\tpi_1(\vz)}{\pi_0(\vz)}\right)\right]^\frac{1}{1 - q}\\
    &= \pi_0(\vz) \, \hl{\exp_q} \left\{ \beta \cdot \ln_q\left(\frac{\tpi_1(\vz)}{\pi_0(\vz)}\right)\right\}. \label{eq:qexp_form}
\end{align}
To mirror the likelihood ratio exponential family interpretation of the geometric path in \cref{eq:lkd_ratio_fam}, we multiply by a factor $Z_q(\beta)$ to write the normalized $q$-path distribution as
\begin{align}
    \pi_{\beta,q}(z) &= \frac{1}{Z_q(\beta)} \, \pi_0(\vz)\exp_q \left\{\beta \cdot \suffq(\vz) \right\} \\ 
    Z_q(\beta) &:= \int \tilde{\pi}_{\beta, q}(\vz) \, d\vz \, , \quad \suffq(\vz) := \ln_q \frac{\tpi_1(\vz)}{\pi_0(\z)}   
\end{align}
which recovers \cref{eq:lkd_ratio_fam} as $q \to 1$.

Note that we normalize using $Z_q(\beta)$ instead of subtracting a $\psi_q(\beta)$ term inside the $\exp_q$ as in the standard definition of a parameteric $q$-exponential family \citep{naudts2009q, naudts2011generalised, amari2011q}
\begin{align}
    \pi_{\theta,q}(\vz) &= g(\vz) \, \exp_q \big\{ \theta \cdot \suffexp_q(\vz) - \psi_q(\theta ) \big\}. \label{eq:qexp_fam1}
\end{align}
where we use $\suffexp_q(\vz)$ to indicate a general sufficient statistic vector which may differ from ${\suffq(\vz) = \ln_{q} \tpi_1(\vz) / \pi_0(\z)}$ above.

While $\log Z(\beta) = \psi(\beta)$ for $q=1$,  translating between these normalization constants for $q \neq 1$ requires a non-linear transformation of the parameters. This delicate issue of normalization has been noted in \citep{matsuzoe2019normalization,suyari2020advantages,naudts2011generalised}, and we give detailed discussion in App. \ref{app:normalization}.
In  App. \ref{sec:student}, we use the $\psi_q(\theta)$ normalization constant to derive an analogue of the moment-averaging path between parametric $q$-exponential family endpoints.

\paragraph{$q$-Paths for Parametric Endpoints}
The geometric path has a particularly simple form when annealing between exponential family endpoint distributions 
\begin{align}
    \theta_{\beta} = (1-\beta) \, \theta_0 + \beta\, \theta_1 \label{eq:geo_path_nat_params} \, .
\end{align}
In \cref{app:same_family}, we verify \cref{eq:geo_path_nat_params} and show that the same result holds for $q$-paths between endpoint distributions within the same $q$-exponential family.
Intuitively, for the (generalized) exponential family distribution in \cref{eq:qexp_fam1}, we can write the unnormalized density ratio $\ln_q \tpi_{\theta}(\vz) / g(\vz) = \theta \cdot \phi(\vz)$ as a linear function of the parameters $\theta$.   Thus, the $q$-path generalized mean over density functions with $h_q(\tpi_{\theta_i}) = \ln_q \tpi_{\theta_i}(\vz)$ will translate to an arithmetic mean in the parameter space with $h_1(\theta_i) = \theta_i$.

\section{Variational Representations} \label{sec:vrep_breg}
\citet{grosse2013annealing} observe that intermediate distributions along the geometric path can be viewed as the solution to a weighted \textsc{kl} divergence minimization
\begin{align}
&\pi_{\beta} = \argmin\limits_{r}(1-\beta)D_{\KL}[r\|\pi_{0}] +\beta D_{\KL}[r\|\pi_{1}] \label{eq:vrep_exp}
\end{align}
where the optimization is over arbitrary distributions $r(\vz)$.

When the endpoints come from an exponential family of distributions and the optimization is limited to only this parametric family $\mathcal{P}_{e}$, \citet{grosse2013annealing} find that the moment-averaged path is the solution to a \textsc{kl} divergence minimization with the order of the arguments reversed
\begin{align}
&\pi_{\eta}=\argmin\limits_{r \in \mathcal{P}_{e}} (1-\beta)D_{\KL}[\pi_{0}\|r]+\beta D_{\KL}[\pi_{1}\|r]\label{eq:vrep_moments}.
\end{align}
In App. \ref{app:alpha_integration}, we follow similar derivations as \citet{amari2007integration} to show that the $q$-path density $\tpi_{\beta,q}$ minimizes the $\alpha$-divergence to the endpoints
\begin{align}
\tpi_{\beta, q}=\argmin\limits_{\tilde{r}}(1- \beta)& D_{\alpha}[\tpi_{0}||\tilde{r}] +\beta D_{\alpha}[\tpi_{1}||\tilde{r}]\label{eq:vrep_alpha}
\end{align}
where the optimization is over arbitrary measures $\tilde{r}(z)$.  Amari's $\alpha$-divergence over unnormalized measures, for $\alpha = 2q-1$ (\cite{amari2016information} Ch. 4),  is defined
\begin{align}
D_{\alpha}[\tilde{r}:\tilde{p} ]& = \small \frac{4}{(1-\alpha^2)} \bigg( \frac{1-\alpha}{2}  \int \tilde{r}(\vz)d\vz \label{eq:alpha_div}\\
&\phantom{=}+ \frac{1+\alpha}{2}  \int \tilde{p}(\vz) d\vz  -\int \tilde{r}(\vz)^{\frac{1-\alpha}{2}} \, \tilde{p}(\vz)^{\frac{1+\alpha}{2}}  d\vz  \bigg) \nonumber
\end{align}
The $\alpha$-divergence variational representation in \cref{eq:vrep_alpha} generalizes \cref{eq:vrep_exp}, since the \textsc{kl} divergence $D_{\KL}[\tilde{r}||\tilde{p}]$ is recovered (with the order of arguments reversed)\footnote{The \textsc{kl} divergence extended to unnormalized measures is defined $D_{KL}[\tilde{q}:\tilde{p} ] = \int \tilde{q}(\vz) \log \frac{\tilde{q}(\vz)}{\tilde{p}(\vz)} d\vz  - \int \tilde{q}(\vz) d\vz +  \int \tilde{p}(\vz) d\vz$.} as $q \rightarrow 1$.

\begin{figure}
  \centering
  \subfigure[\text{Moment-Avg}]{\includegraphics[trim={0 0 0 0},clip, scale=.25]{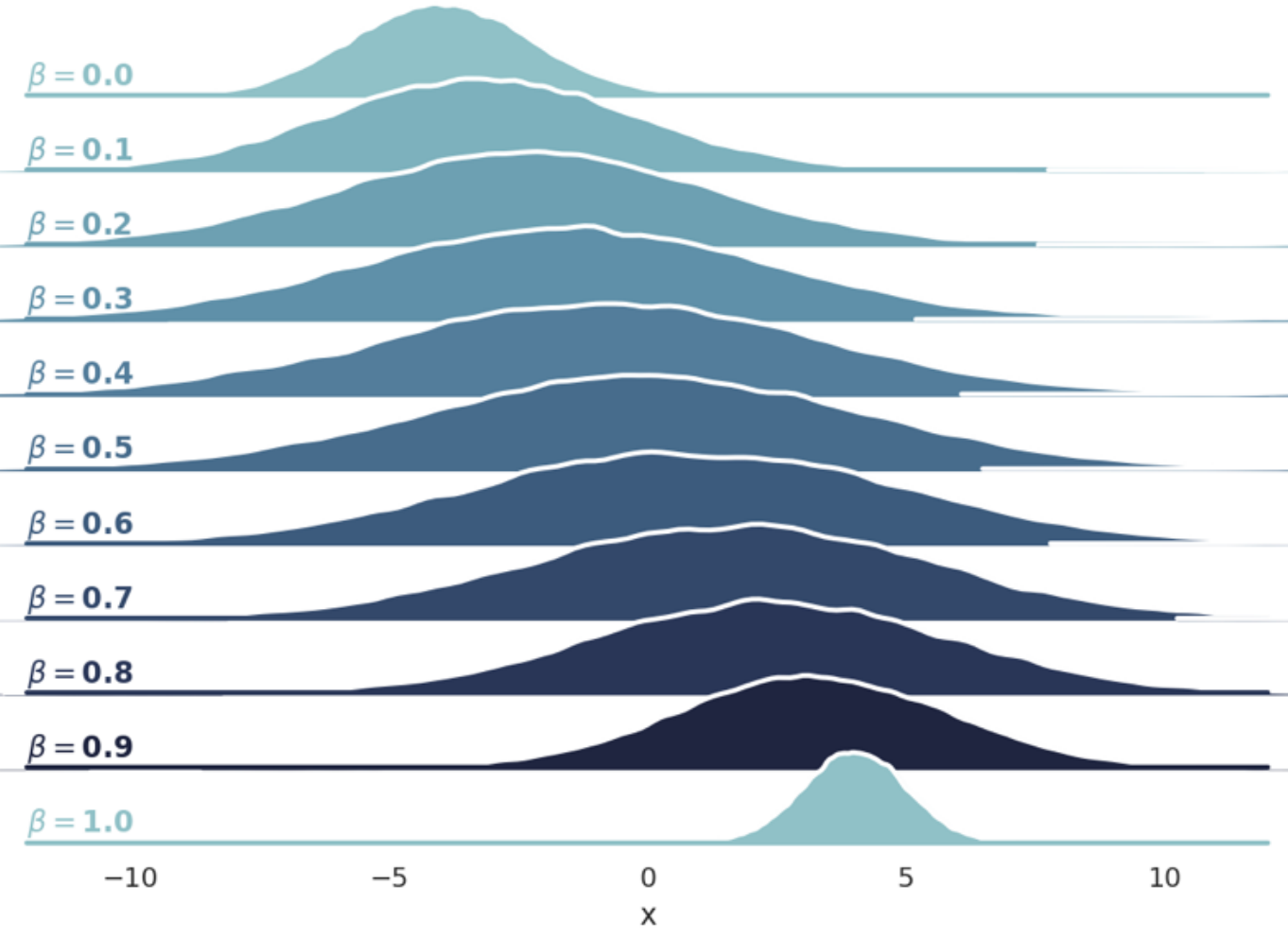}}
  \subfigure[$q=0$]{\includegraphics[trim={0 0 0 0},clip, scale =.25]{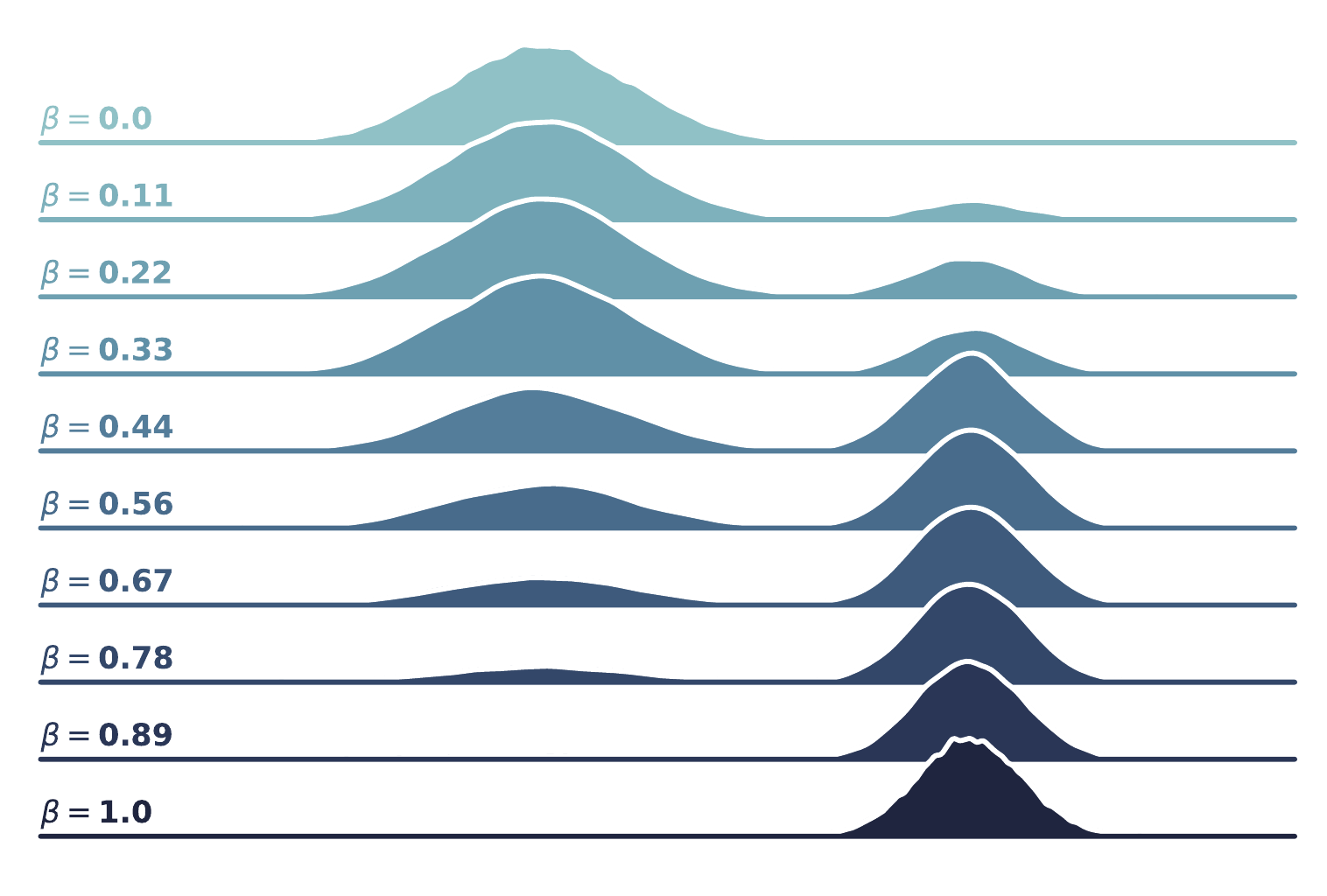}}
  \caption{Moment-averaging path and $q=0$ mixture path between $\mathcal{N}(-4, 3)$ and $\mathcal{N}(4,1)$. See \cref{sec:vrep_breg}, \cref{app:mixture_path}, and \cref{app:moments_as_generalized_mean} for discussion.}
  \label{fig:moments_vs_mixture}
\end{figure}

However,  while the $\alpha$-divergence tends to $D_{\KL}[\tilde{p}||\tilde{r}]$ as ${q \to 0}$, \cref{eq:vrep_alpha} \textit{does not} generalize \cref{eq:vrep_moments} since the optimization in \cref{eq:vrep_moments} is restricted to the parametric family $\mathcal{P}_{e}$.
For the case of arbitrary endpoints, the \textit{mixture} distribution rather than the moment-averaging distribution minimizes the reverse \textsc{kl} divergence in \cref{eq:vrep_moments}, producing different paths as seen in \cref{fig:moments_vs_mixture}.  We discuss this distinction in greater detail in \cref{app:mixture_path} and \cref{app:moments_as_generalized_mean}.




\section{Related Work}\label{sec:related}
In \cref{sec:path_exp_fam} and \cref{app:parametric}, we discuss connections between  $q$-paths and the $q$-exponential family.   Examples of parametric $q$-exponential families  include the Student-$t$ distribution, which has the same first- and second-moment sufficient statistics as the Gaussian and a degrees of freedom parameter $\nu$ that specifies a value of $q > 1$.   This induces heavier tails than the standard Gaussian and leads to conjugate Bayesian interpretations in hypothesis testing with finite samples \citep{murphy2007conjugate, gelman2013bayesian}.
The generalized Pareto distribution is another member of the $q$-exponential family, and has been used for modeling heavy-tail behavior \citep{pickands1975statistical, bercher2008new, tsallis2009introduction}, smoothing outliers for importance sampling estimators \citep{vehtari2015pareto}, or evaluating variational inference \citep{yao2018yes}.  
$q$-logarithms and exponentials have also appeared in methods for classification \citep{ding2011t, amid2019two}, robust hypothesis testing \citep{qin2017robust}, mixture modeling \citep{qin2013maximum}, variational inference \citep{ding2011t, kobayashi2020q}, and expectation propagation \citep{futami2017expectation, minka2004power}.

In \cref{sec:vrep_breg}, we showed that each $q$-path density $\tpi_{\beta,q}(\vz)$ specifies the minimizing argument for a variational objective in \cref{eq:vrep_exp} or \cref{eq:vrep_alpha}.  The value of the objective in \cref{eq:vrep_exp} is a mixture of \textsc{kl} divergences, and can be interpreted as a generalized Jensen-Shannon divergence \citep{Nielsen_2019, nielsen2021variational} or Bregman information \citep{Banerjee2005}.   \citet{deasy2021constraining} explore this mixture of divergences as a regularizer in variational inference, while \citet{brekelmans2020lref} provide additional analysis for case of $q=1$. 


 \begin{figure}[t]
            \begin{minipage}{\columnwidth}
            \captionof{table}{SMC sampling with linear/adaptive scheduling in a binary regression model for $\{1,3,5\}$ move steps. \textsc{lin} indicates a linearly spaced schedule ($K=10$) and \textsc{ada} uses an adaptive schedule (cf. \cref{sec:experiments_smc}).  Median $\textsc{err} = |\log\hat{p}(D) - \log p(D)|$ across $10$ seeds is reported against ground truth. \textsc{q-path (grid)} shows best of 20 log-spaced $\delta \in [10^{-5}, 10^{-1}]$, and \textsc{q-path (ess)} uses the $\textsc{ess}$ heuristic to initialize $q$ as described in \ref{app:smc_exp_details}.  Error for most runs (8/12) is {\textsc{q-path (grid)} < \textsc{q-path (ess)} < \textsc{geo}}. }
        \label{tab:pima_table}
        \centering
        \scalebox{0.90}{
        \begin{tabular}{lrrr}
            \toprule
             &     &  \multicolumn{1}{c}{\textsc{q-path}}  &  \multicolumn{1}{c}{\textsc{q-path}} \\
             \multicolumn{1}{l}{\textsc{pima}}&   \multicolumn{1}{c}{\textsc{geo}}  &  \multicolumn{1}{c}{\small \textsc{(ess heuristic)}}  &  \multicolumn{1}{c}{\small \textsc{(grid)}} \\
            \midrule
            \textsc{lin-1}   & 79.02 (39.1)  & 80.64 (42.33)  & \textbf{10.77 (2.30)}\\
            \textsc{lin-3}   & 59.11 (41.71)  & 59.64 (47.41)  & \textbf{5.79 (1.46)}\\
            \textsc{lin-5}   & 45.63 (19.86)  & 41.96 (25.23)  & \textbf{6.63 (2.62)}\\
            \textsc{ada-1} &  2.51 (1.35)  & 2.31 (2.99)  & \textbf{1.62 (1.79)}\\
            \textsc{ada-3} &  1.49 (0.43)  & 1.12 (1.05)  & \textbf{0.84 (0.84)}\\
            \textsc{ada-5} &  \textbf{0.48 (0.60)}   & 0.76 (0.29)  & 0.52 (0.59)\\[.5ex]
            \multicolumn{1}{l}{\textsc{sonar}}&\multicolumn{2}{l}{} & \\ \midrule
            \textsc{lin-1}   & 228.7 (80.9)             & 217.92 (72.51)            & \textbf{93.33 (15.79)}\\
            \textsc{lin-3}   & 175.21 (38.66)           & 172.66 (61.55)            & \textbf{55.94 (5.69)}\\
            \textsc{lin-5}   & 218.94 (92.08)           & 222.07 (78.76)            & \textbf{36.67 (10.32)}\\
            \textsc{ada-1} & 20.17 (15.99)            & 18.15 (15.43)            & \textbf{15.32 (8.19)}\\
            \textsc{ada-3} & 3.83 (3.44)             & 3.78 (2.77)           &  \textbf{3.11 (3.26)}\\
            \textsc{ada-5} & 2.79 (2.41)             & 2.68 (1.95)           &  \textbf{2.23 (0.72)}\\
             \bottomrule
        \end{tabular}}
\end{minipage}\hspace*{.075\columnwidth}
\end{figure}

\section{Experiments}\label{sec:experiments}
Code for all experiments is available at \url{https://github.com/vmasrani/qpaths_uai_2021}.

\subsection{Sequential Monte Carlo in Bayesian Inference}\label{sec:experiments_smc}

\begin{figure}[t]
\begin{minipage}{.95\columnwidth}
    \vspace*{-.1cm}
    \centering
    \includegraphics[width=0.9\textwidth]{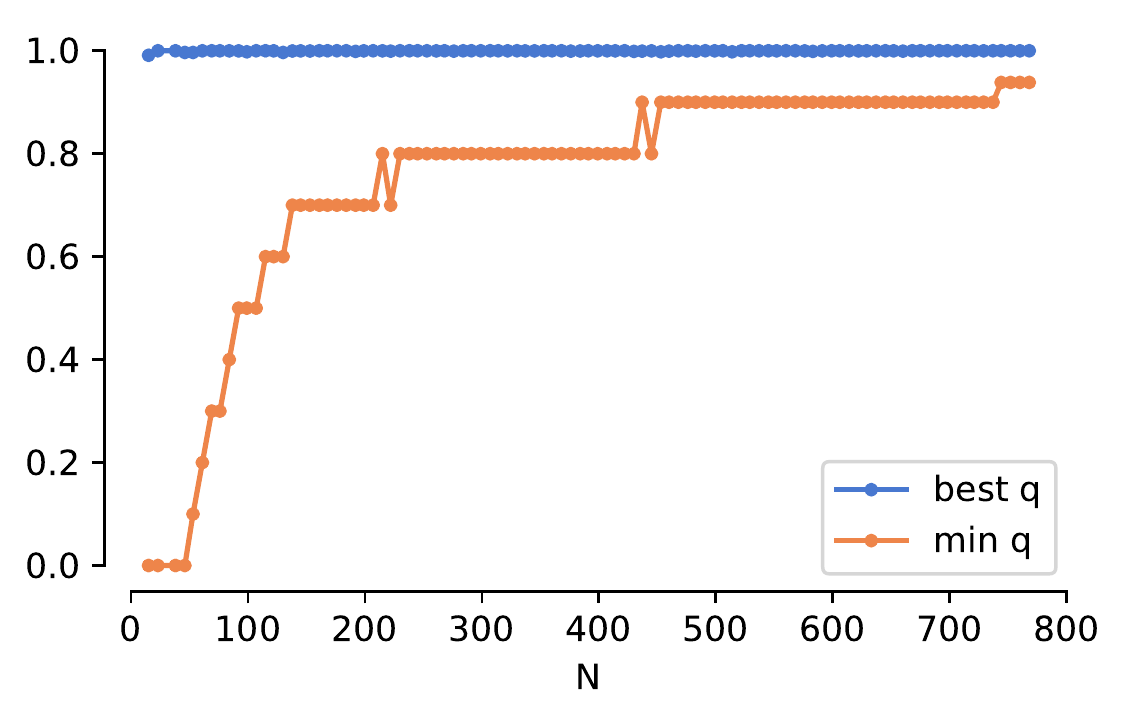} 
    \vspace*{-0.1cm}
    \captionof{figure}{Evaluating the choice of $q$ for \gls{SMC}.  Since the scale of the likelihood $\tpi_1$ depends on the number of data examples, we expect the numerical stability of $q$-paths to vary by $N$.  While the minimum $q$ yielding a stable estimator (orange) increases with $N$,
    the best performing $q$-path (blue) is still $q=1-\delta$ for small $\delta >0$.
}\label{fig:varying_n}
\end{minipage}
\end{figure}

In this section, we use \gls{SMC} to sample posterior parameters $\pi_1(\theta) = p(\theta|\mathcal{D}) \propto p(\theta) \prod_{n=1}^N p(\vx_n | \theta)$ and estimate the log marginal likelihood $\log p(\mathcal{D})= \log \int p(\theta)p(\mathcal{D}|\theta) d\theta$ in a Bayesian logistic regression models on the ``tall'' Pima Indians diabetes dataset ($N=768, D=8$) and ``wide'' Sonar dataset ($N=208, D=61$) (see \cref{sec:experiment_details}).  Ground truth $\log p(D)$ is computed using 50k samples and 20 move steps, and for all runs we use 10k samples and plot median error across ten seeds. Grid search shows best of 20 runs, where we sweep over 20 log-spaced $\delta \in [10^{-5}, 10^{-1}]$.

We explore the use of $q$-paths in both the non-adaptive case, with a fixed linear $\beta$ schedule with $K=10$ intermediate distributions, and the adaptive case, where the next value of $\beta_{t+1}$ is chosen to yield an \gls{ESS} of $N/2$ \citep{chopin2020introduction}. 

For the non-adaptive case, we find in \cref{fig:smc_sampling} that $q \in [0.9954, 0.9983]$ can achieve more accurate marginal likelihood estimates than the geometric path with fewer movement steps and drastically reduced variance. In \cref{tab:pima_table} we see that $q$-paths achieve gains over the geometric path in both the linear and adaptive setting across both datasets.
\begin{figure*}[!t]
    \begin{minipage}{.49\textwidth}
    \vspace*{-.15cm}
    \centering
    \includegraphics[width=0.99\columnwidth]{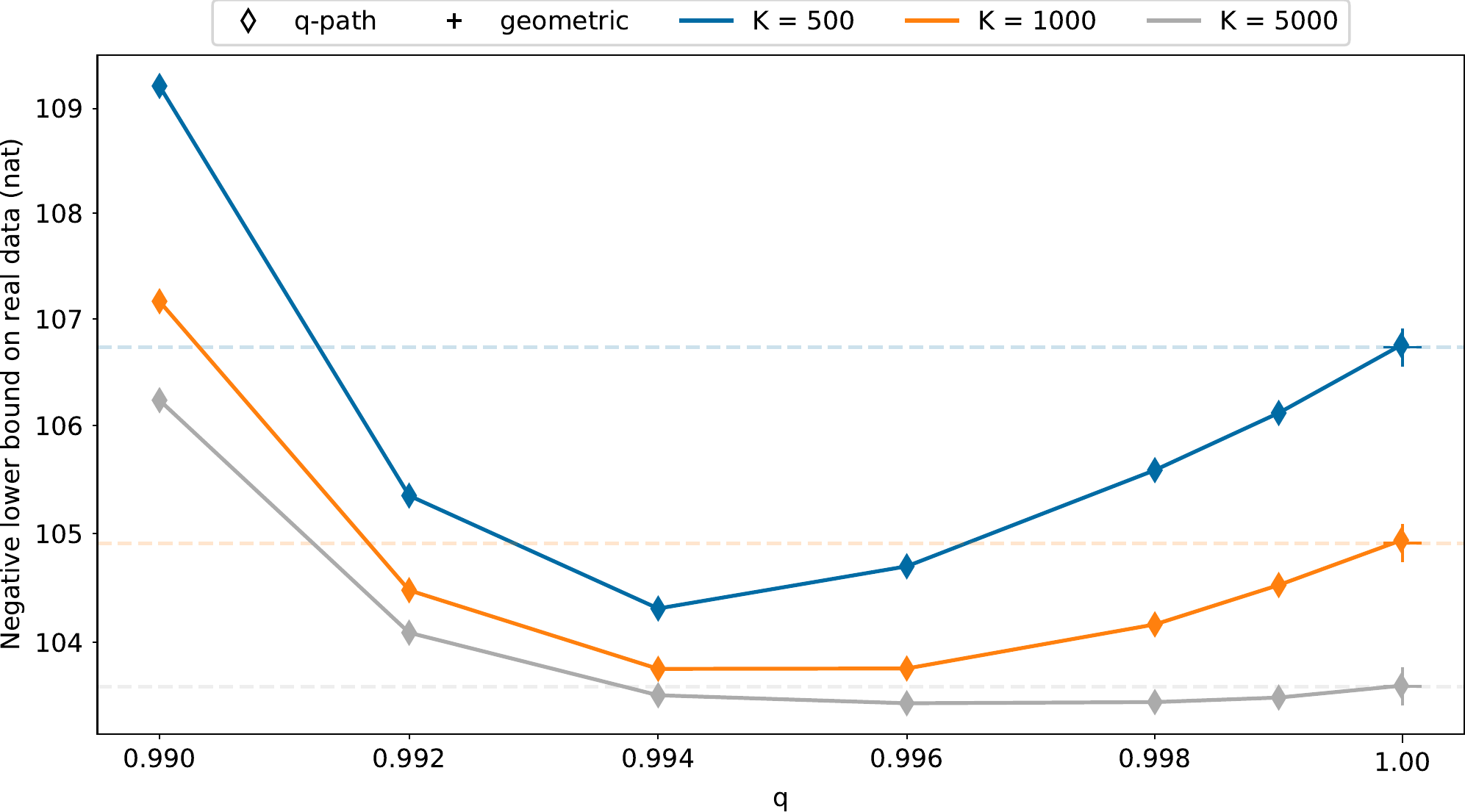}
    \vspace*{-.05cm}
    \captionof{subfigure}{Estimating $\log p(x)$ on real data using \gls{AIS}.}\label{fig:bdmc_omniglot_real}
    \end{minipage}
    \begin{minipage}{.49\textwidth}
    \vspace*{-.1cm}
    \subfigure{\includegraphics[width=0.99\columnwidth]{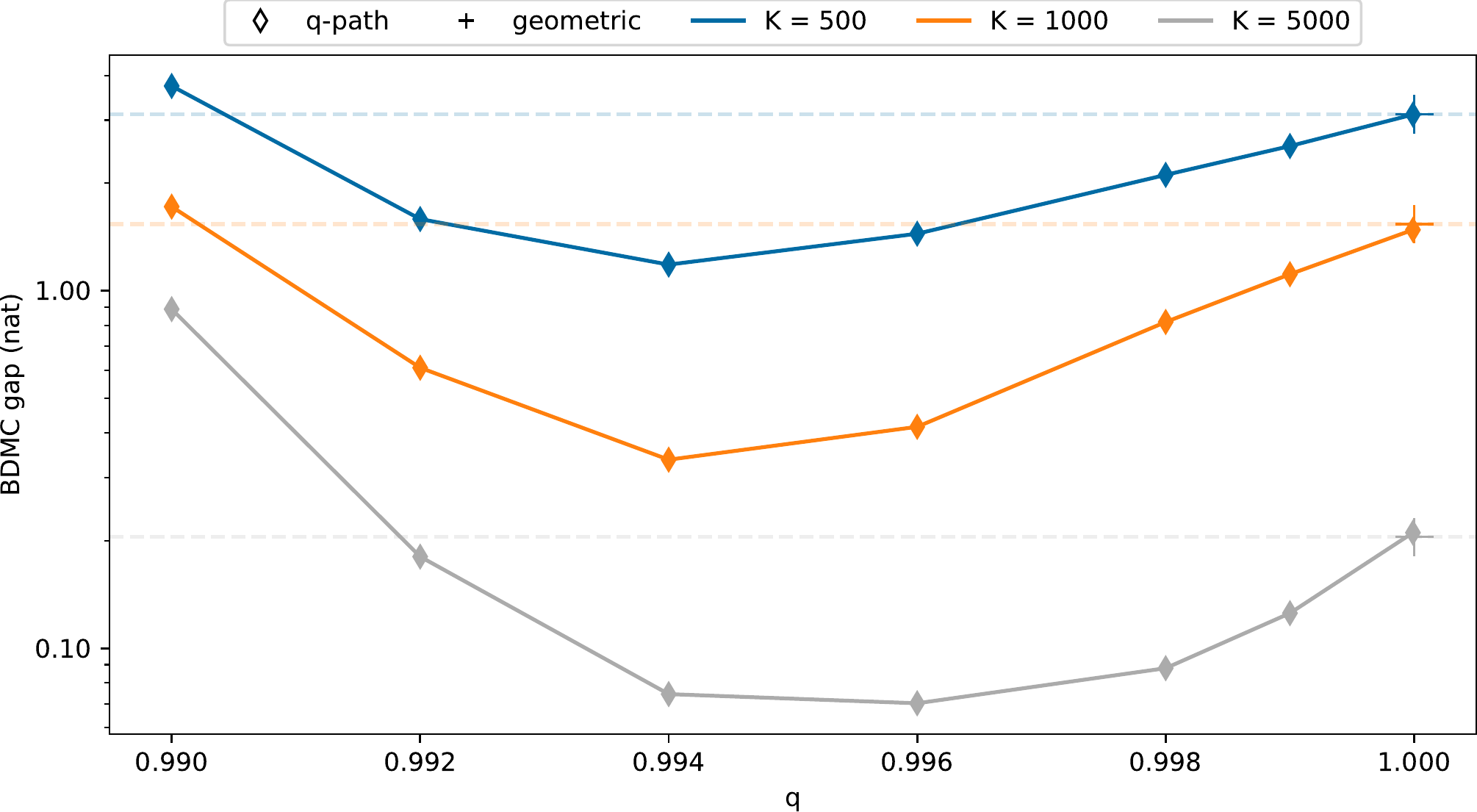}}
    \vspace*{-.25cm}
    \captionof{subfigure}{\acs{BDMC} Gap on simulated data.}\label{fig:bdmc_omniglot_gap}
    \end{minipage}
    \vspace*{-.15cm}
    \caption{Evaluating Generative Models using \gls{AIS} with $q$-paths on Omniglot dataset.  Best viewed in color.}
    \label{fig:bdmc_omniglot_result}
    \end{figure*}

\paragraph{Numerical Stability and Implementation}

To implement $q$-paths in practice, we begin by considering the log of the expression in \cref{eq:qexp_form}, which is guaranteed to be non-negative because $\tilde{\pi}_{\beta, q}(\vz)$ is an unnormalized density.
 \begin{align}
     &\log \tilde{\pi}_{\beta, q}(\vz) = \\
     &\log \pi_0(\vz) + \frac{1}{1 - q} \log \left[1 + (1-q) \cdot \beta \cdot \ln_q\left( \frac{\tpi_1(\vz)}{\pi_0(\vz)}\right)\right]\nonumber, \label{eq:q_energy}
 \end{align}
We focus attention on  $\ln_q \tpi_1(\vz)/\pi_0(\vz)$ term, which is potentially unstable for $q\neq 1$ since it takes importance weights $w =\tpi_1(\vz)/\pi_0(\vz)$ as input.
Since we are usually given log weights in practice, we consider the identity mapping $w = \exp (\log w)$ and reparameterize ${q = 1 - \frac{1}{\rho}}$ to obtain
\begin{align}
    \ln_q\left(\exp \log w\right) &= \frac{1}{1-q}\left[ \left(\exp \log w \right)^{1-q} - 1\right]\\
                             &= \rho\left[ \left(\exp \log w \right)^{\frac{1}{\rho}} - 1\right]\\
                              &= \rho\left[\exp \{ \frac{1}{\rho} \log w \} - 1\right] \, .
\end{align}

This suggests $q$ should be chosen such that the exponential doesn't overflow or underflow, which can be accomplished by setting $\rho$ on the order of
\begin{align}
    \rho = \max_i |\log w_i|.  \label{eq:rho_choice}
\end{align}
where $i$ indexes a set of particles $\{\vz_i\}$.
This choice is reminiscent of the log-sum-exp trick and ensures $|\frac{1}{\rho} \log w| \le 1$.

In \cref{fig:varying_n}, we explore the impact of changing the scale of $\log w$ on the numerical stability of $q$-paths.   For the case of inferring global model parameters over $N$ i.i.d. data points $p(\mathcal{D}) = \prod_{n=1}^N p(\vx_n)$, we can see that
 the scale of the unnormalized densities $\tpi_1(\theta, \mathcal{D}) = p(\theta) \prod_{n=1}^N p(\vx_n | \theta)$ differs based on the number of datapoints, where increasing $N$ decreases the magnitude of $\log w = \log \tpi_1(\theta, \mathcal{D})$ with $\tpi_0(\theta) = p(\theta)$.

 We randomly subsample $N$ data points for conditioning our model, and observe the effect on both the best-performing $q$ and the numerical stability of \gls{SMC} with $q$-paths. The minimum value of $q$ for which we can obtain stable estimators rises as the number of datapoints $N$ increases and the scale of $\tpi_1(\theta, \mathcal{D})$ becomes smaller.  

\paragraph{Sensitivity to $q$}
While setting $\rho$ on the order of $\max_i |\log w_i|$ ensures numeric stability,  \cref{fig:varying_n} indicates that numerical stability may not be sufficient for achieving strong performance in \gls{SMC}.   In fact, $q$-paths with values just less than $1$ consistently perform best across all values of $N$.

To understand this observation, recall the example in Fig. \ref{fig:q_path_separated} where the initial and target distribution are well-separated and even the $q=0.98$ path begins to resemble a mixture distribution.   This is clearly undesirable for path sampling techniques, where the goal is to bridge between base and target densities with distributions that are easier to sample.

\paragraph{Heuristic for Choosing $q$}
Motivated by the observations above and the desire to avoid grid search, we provide a rough heuristic to find a $q$ which is well-suited to a given estimation problem.

Taking inspiration from the \gls{ESS} criterion used to select $\beta_{t+1}$ in our \gls{SMC} experiments above   \citep{chopin2020introduction}, we select $q$ to obtain a target value of \gls{ESS} for the first intermediate $\beta_1$
\begin{align}
    \mathcal{L}(\beta_1, q) &= ||\text{ESS}(\beta_1, q) - \text{ESS}_\text{target} ||_2^2 \label{eq:ess_loss} \\
    \text{ESS}(\beta, q) &= \frac{\big(\sum_{i} w_i(\beta, q)\big)^2 }{\sum_i w_i\big(\beta, q\big)^2} \, \, \text{with} \, \, w_i(\beta, q) = \frac{\tpi_{\beta, q}(z_i)}{\pi_0(z_i)}. \nonumber
\end{align}
As in the case of the adaptive $\beta$ scheduling heuristic for \gls{SMC}, we set the target $\text{ESS}_\text{target} = N/2$ to ensure adequate sampling diversity \citep{jasraInferenceLevyDrivenStochastic2011, schaferSequentialMonteCarlo2013,buchholzAdaptiveTuningHamiltonian2021, chopin2020introduction}.
For fixed scheduling, the value of $\beta_1$ may be known and thus we can easily select $q$ to obtain the target value $\text{ESS}(\beta_1, q) \approx \text{ESS}_\text{target}$.    However, in adaptive scheduling, $\beta_1$ is not known and the objective $\mathcal{L}(\beta_1, q)$ is non-convex in $\beta_1, q$.   In \cref{app:ais_exp_details}, we provide a coordinate descent algorithm to find local optima using random initializations around an initial $q = 1- \frac{1}{\rho}$ for $\rho$ as in \cref{eq:rho_choice}, with results in \cref{tab:pima_table}.

Note that this heuristic sets $q$ based on a set of initial $z_i \sim \pi_0(\vz)$, and thus does not consider information about the \gls{MCMC} sampling used to transform and improve samples.

Nevertheless, in \cref{tab:pima_table} we observe that $q$-paths initialized by this heuristic can outperform the geometric path on benchmark \gls{SMC} binary regression tasks.   
Comparison with grid search results indicate that further performance gains might be achieved with an improved heuristic.




\subsection{Evaluating generative models using AIS} \label{sec:experiments_ais}
\gls{AIS} with geometric paths is often considered the gold-standard for evaluating decoder-based generative models \citep{wuetal17}.   In this section, we evaluate whether $q$-paths can improve marginal likelihood estimation for a \gls{VAE} trained using the \gls{TVO} \citep{masrani2019thermodynamic} on the Omniglot dataset.

First, we use \gls{AIS} to evaluate the trained generative model on the true test set, with a Gaussian prior $\pi_0(\vz) = p(\vz)$  as the base distribution and true posterior $\pi_1(\vz) = p(\vz|\vx) \propto p(\vx,\vz)$ as the target.   Intermediate distributions then become $\tpi_{\beta}(\vz) = p(\vz)p(\vx|\vz)^{\beta}$.   We report stochastic lower bound estimates \citep{grosse2015sandwiching} of $\mathbb{E}_{p_{\text{data}}(x)} \log p(x)$ in \cref{fig:bdmc_omniglot_gap}, where we have plotted the negative likelihood bound so that lower is better.  Even for a large number of intermediate distributions, we find that $q \in [0.992, 0.998]$ can outperform the geometric path.


When exact posterior samples are available, we can use a reverse \gls{AIS} chain from the target density to the base to obtain a stochastic {\it upper bound} on the $\log$ marginal likelihood \citep{grosse2015sandwiching}.  While such samples are not available on the real data, we can use simulated data drawn from the model using ancestral sampling $x, \vz \sim  p(z)p(x|z)$ as the dataset, and interpret $\vz$ as a posterior sample.    We use the \gls{BDMC} gap, or difference between the stochastic lower and upper bounds obtained from forward and reverse chains on simulated data, to evaluate the quality of the \gls{AIS} procedure.


In \cref{fig:bdmc_omniglot_result}, we report the average \gls{BDMC} gap on $2500$ simulated data examples, and observe that $q$-paths with $q=0.994$ or $q=0.996$ consistently outperform the geometric path as we vary the number of intermediate distributions $K$.

\section{Conclusion}
In this work, we proposed $q$-paths as a generalization of the geometric mixture path which can be constructed between arbitrary endpoint distributions and admits a closed form energy function.   We provided a $q$-likelihood ratio exponential family interpretation of our paths, and derived a variational representation of $q$-path intermediate densities as minimizing the expected $\alpha$-divergence to the endpoints.   
Finally, we observed empirical gains in \gls{SMC} and \gls{AIS} sampling using $q$-paths with $q = 1-\delta$ for small $\delta$.   

Future work might consider more involved heuristics for choosing $q$, such as running truncated, parallel sampling chains, to capture the interplay between choices of $\beta, q,$ and sampling method.  
Applying $q$-paths in settings such as sampling with \gls{PT} or variational inference using the \gls{TVO}, remain interesting questions for future work.






\begin{acknowledgements}
RB and GV acknowledge support from the Defense Advanced Research Projects Agency (DARPA) under awards
FA8750-17-C-0106 and W911NF-16-1-0575.
VM and FW acknowledge the support of the Natural Sciences and Engineering Research Council of Canada (NSERC), the Canada CIFAR AI Chairs Program, and the Intel Parallel Computing Centers program. This material is based upon work supported by the United States Air Force Research Laboratory (AFRL) under the Defense Advanced Research Projects Agency (DARPA) Data Driven Discovery Models (D3M) program (Contract No. FA8750-19-2-0222) and Learning with Less Labels (LwLL) program (Contract No.FA8750-19-C-0515). Additional support was provided by UBC's Composites Research Network (CRN), Data Science Institute (DSI) and Support for Teams to Advance Interdisciplinary Research (STAIR) Grants. This research was enabled in part by technical support and computational resources provided by WestGrid (\url{https://www.westgrid.ca/}) and Compute Canada (\url{www.computecanada.ca}).

\end{acknowledgements}
\bibliographystyle{plainnat}
\bibliography{masrani_722}
\clearpage
\onecolumn
\appendix
\section{Abstract Mean is Invariant to Affine Transformations} \label{app:any_h}
In this section, we show that $h_{q}(u)$ is invariant to affine transformations. That is, for any choice of $a$ and $b$,
\begin{align}
    h_{q}(u) =
\begin{cases}
    a \cdot u^{1-q} + b \hfill & q \neq 1 \\
    \log u \hfill 	&  q = 1
\end{cases} \label{eq:alpha_abstract2}
\end{align}
yields the same expression for the abstract mean $\mu_{h_{\alpha}}$. First, we note the expression for the inverse $h^{-1}_{q}(u)$ at $q \neq 1$
\begin{align}
    h^{-1}_{q}(u) = \left(\frac{u - b}{a}\right)^{\frac{1}{1-q}}.
\end{align}
Recalling that $\sum_i w_i = 1$, the abstract mean then becomes
\begin{align}
    \mu_{h_{q}}(\{w_i\}, \{u_i\}) &= h_{q}^{-1}\left(\sum_i w_i h_{q}(u_i) \right) \\
    &= h_{q}^{-1}\left(a \left(\sum_i w_iu_i^{1-q}\right) + b \right) \\
    &=\bigg(\sum_i w_i u_i^{1-q} \bigg)^{\frac{1}{1-q}}
\end{align}
which is independent of both $a$ and $b$.

\section{Normalization in q-Exponential Families}\label{app:normalization}
The $q$-exponential family can also be written using the $q$-free energy $\psi_q(\theta)$ for normalization \cite{amari2011q, naudts2011generalised},
\begin{align}
    \pi_{\theta,q}(\z)  &= \pi_0(z) \, \exp_q \big\{ \theta \cdot \phi(\vz) - \psi_q(\theta ) \big\} \, . \label{eq:qexp_fam_qf} 
\end{align}
However, since $\exp_q \{ x + y \} = \exp_q \{ y \} \cdot \exp_q \{ \frac{x}{1 + (1-q) y}  \}$ (see \cite{suyari2020advantages} or App. \ref{app:q_sum_product} below) instead of $\exp \{ x + y \} = \exp \{ x \} \cdot \exp \{ y \} $ for the standard exponential, we can not easily move between these ways of writing the $q$-family \cite{matsuzoe2019normalization}.

Mirroring the derivations of \citet{naudts2011generalised} pg. 108, we can rewrite \eqref{eq:qexp_fam_qf} using the above identity for $\exp_q \{ x+y\}$, as
\begin{align}
\pi^{(q)}_{\theta}(\vz) &= \pi_0(\vz) \, \exp_q \{ \theta \cdot \phi(\vz) - \psi_q(\theta) \} \label{eq:normalization1} \\
&= \pi_0(\vz) \, \exp_q \{ - \psi_q(\theta) \} \exp_q \big\{ \frac{\theta \cdot \phi(\vz)}{1+(1-q)(-\psi_q(\theta))} \big \} \label{eq:normalization2}
\end{align}
Our goal is to express $\pi^{(q)}_{\theta}(\vz)$ using a normalization constant $Z^{(q)}_\beta$ instead of the $q$-free energy $\psi_q(\theta)$.  While the exponential family allows us to freely move between $\psi(\theta)$ and $\log Z_{\theta}$, we must adjust the natural parameters (from $\theta$ to $\beta$) in the $q$-exponential case.   Defining
\begin{align}
 \beta &= \frac{\theta}{1+(1-q)(-\psi_q(\theta))} \\
Z^{(q)}_\beta &= \frac{1}{\exp_q \{-\psi_q(\theta) \}} 
\end{align}
we can obtain a new parameterization of the $q$-exponential family, using parameters $\beta$ and multiplicative normalization constant $Z^{(q)}_\beta$,
\begin{align}
 \pi_{\beta,q}(\vz) &=  \frac{1}{Z^{(q)}_\beta} \pi_0(z) \, \exp_{q} \{ \beta \cdot \phi(\vz) \} \\
 &= \pi_0(z) \, \exp_q \big\{ \theta \cdot \phi(\vz) - \psi_q(\theta ) \big\} = \pi^{(q)}_{\theta}(\vz)  \, .
\end{align}
See \citet{matsuzoe2019normalization}, \citet{suyari2020advantages}, and \citet{naudts2011generalised} for more detailed discussion of normalization in deformed exponential families.

\section{Minimizing $\alpha$-divergences}\label{app:alpha_integration}
\citet{amari2007integration} shows that the $\alpha$ power mean $\pi^{(\alpha)}_{\beta}$ minimizes the expected divergence to a single distribution, for \textit{normalized} measures and $\alpha = 2q-1$.   We repeat similar derivations for the case of unnormalized endpoints $\{\tpi_i\}$ and $\tilde{r}(\vz)$ and show
\begin{align}
    \tpi_{\beta, q} = \argmin \limits_{\tilde{r}(z)}(1-\beta)& D_{\alpha}[\tpi_{0}(z)||\tilde{r}(z)] +\beta D_{\alpha}[\tpi_{1}(z)||\tilde{r}(z)],
\end{align}
for $\alpha = 2q-1$.
\begin{proof}
    Defining $w_0 = (1 - \beta)$ and $w_1 = \beta$, we consider minimizing the functional
    \begin{align}
        r^*(z) &=\argmin_{\tilde{r}(z)} J[r(z)]
        =\argmin_{\tilde{r}(z)} \left(\sum_{i=0}^{N=1} w_i D_\alpha(\tpi_i(\vz)||\tilde{r}(\vz))  \right)\label{eq:amari_lagrange}
    \end{align}
    \cref{eq:amari_lagrange} can be minimized using the Euler-Lagrange equations or using the identity
    \begin{align}
        \frac{\delta f(x)}{\delta f(x')} = \delta(x - x')\label{eq:delta_eq}
    \end{align}
    from \cite{meng2004introduction}. We compute the functional derivative of $J[r(z)]$ using \eqref{eq:delta_eq}, set to zero and solve for $r$:
    \begin{align}
        \frac{\delta J[r(z')]}{\delta r(z)} &= \frac{\delta}{\delta r(z)} \left(\sum_{i=0}^{N=1} w_i \left(\frac{1}{q} \int\tilde{p}(z')dz + \frac{1}{1-q} \int\tilde{r}(z')dz - \frac{1}{q(q-1)} \int {\tpi_i(z')}^{1-q}r(z')^{q} dz' \right)  \right)\\
        &= \left(\sum_{i=0}^{N=1} w_i \left(\frac{1}{1-q} \int \hl{\frac{\delta \tilde{r}(z')}{\delta r(z)}} dz - \frac{1}{q(q-1)} \int {\tpi_i(z')}^{1-q} \cdot \hl{q}\cdot r(z')^{\hl{q-1}} \hl{\frac{\delta \tilde{r}(z')}{\delta r(z)}}dz' \right)  \right)\\
        &= \left(\sum_{i=0}^{N=1} w_i \left(\frac{1}{1-q} \int \hl{\delta(z - z')} dz - \frac{1}{q-1} \int {\tpi_i(z')}^{1-q} \cdot r(z')^{\hl{q-1}} \hl{\delta(z - z')}dz' \right)  \right)\\
        0 &= \frac{1}{1-q}\sum_{i=0}^{N=1} w_i \left(1 - {\tpi_i(z)}^{1-q} \cdot r(z)^{q-1} \right) \\
        \sum_{i=0}^{N=1} w_i &=  \sum_{i=0}^{N=1} w_i {\tpi_i(z)}^{1-q} \cdot r(z)^{q-1} \\
        1 &=  \sum_{i=0}^{N=1} w_i {\tpi_i(z)}^{1-q} \cdot r(z)^{q-1} \\
        r(z)^{1-q}  &=  \sum_{i=0}^{N=1} w_i {\tpi_i(z)}^{1-q} \\
        r(z)        &=  \left[(1-\beta) {\tpi_0(z)}^{1-q}  + \beta{\tpi_1(z)}^{1-q}\right]^{1/1-q} =  \tpi_{\beta,q}(z)
    \end{align}
\end{proof}

This result is similar to a general result about Bregman divergences in \citet{Banerjee2005} Prop. 1. although $D_{\alpha}$ is not a Bregman divergence over normalized distributions.

\subsection{Arithmetic Mean ($q=0$)}\label{app:mixture_path}
\newcommand{\pa}{\pi_0}
\newcommand{\pb}{\pi_1}
\newcommand{\opt}{r}
For normalized distributions, we note that the moment-averaging path from \citet{grosse2013annealing} is not a special case of the $\alpha$-integration \cite{amari2007integration}.   While both minimize a convex combination of reverse \textsc{kl} divergences, \citet{grosse2013annealing} minimize within the constrained space of exponential families, 
 while \citet{amari2007integration} optimizes over all normalized distributions.

More formally, consider minimizing the functional
\begin{align}
    J[\opt] &= (1-\beta)D_{\KL}[\pi_{0}(z)||r(z)] +\beta D_{\KL}[\pi_{1}(z)||r(z)] \\
    &= (1 - \beta)\int \pa(z) \log \frac{\pa(z)}{\opt(z)} dz + \beta \int \pb(z) \log \frac{\pb(z)}{\opt(z)} dz \\
            &= \text{const} - \int \big[(1 - \beta) \pa(z) + \beta \pb(z) \big] \cdot \log \opt(z) dz \label{eq:functional}
\end{align}
We will show how \citet{grosse2013annealing} and \citet{amari2007integration} minimize \eqref{eq:functional}.

\paragraph{Solution within Exponential Family}
\citet{grosse2013annealing} constrains $\opt(z) = \frac{1}{Z(\theta)} h(z) \exp (\theta^T g(z))$ to be a (minimal) exponential family model and minimizes \eqref{eq:functional} w.r.t $\opt$'s  natural parameters $\theta$ (cf. \cite{grosse2013annealing} Appendix 2.2):
\begin{align}
    \theta^*_i &= \argmin_\theta J(\theta) \\
    &= \argmin_\theta \left(- \int \big[(1 - \beta) \pa(z) + \beta \pb(z) \big] \left[ \log h(z) + \theta^T g(z) - \log Z(\theta) \right] dz \right)\\
    &= \argmin_\theta \left(\log Z(\theta) - \int \big[(1 - \beta) \pa(z) + \beta \pb(z) \big] \theta^T g(z) dz  + \text{const}\right)
\end{align}
where the last line follows because $\pa(z)$ and $\pb(z)$ are assumed to be correctly normalized. Then to arrive at the moment averaging path, we compute the partials $\frac{\partial J(\theta)}{\partial \theta_i}$ and set to zero:
\begin{align}
    \frac{\partial J(\theta)}{\partial \theta_i} &= \E_{\opt}[g_i(z)] - (1 - \beta)\E_{\pa}[g_i(z)] - \beta \E_{\pb}[g_i(z)] = 0 \\
    \E_{\opt}[g_i(z)] &= (1 - \beta)\E_{\pa}[g_i(z)] - \beta \E_{\pb}[g_i(z)]
\end{align}
where we have used the exponential family identity $\frac{\partial \log Z(\theta)}{\partial \theta_i} = \E_{\opt_{\theta}}[g_i(z)]$ in the first line.

\paragraph{General Solution}
Instead of optimizing in the space of minimal exponential families, \citet{amari2007integration} instead adds a Lagrange multiplier to \eqref{eq:functional} and optimizes $\opt$ directly (cf. \cite{amari2007integration} eq. 5.1 - 5.12)
\begin{align}
    {\opt}^* &= \argmin_{\opt} J'[\opt] \\
    &= \argmin_{\opt} J[\opt] + \lambda \left(1 - \int \opt(z) dz\right) \label{eq:lagrange}
\end{align}
We compute the functional derivative of $J'[\opt]$ using \eqref{eq:delta_eq} and solve for $r$:
\begin{align}
    \frac{\delta J'[\opt]}{\delta \opt(z)}=&- \int \big[(1 - \beta) \pa(z') + \beta \pb(z') \big] \frac{1}{\opt(z')} \frac{\delta \opt(z')}{\delta \opt(z)} dz' - \lambda \int \frac{\delta \opt(z')}{\delta \opt(z)} dz' \\
    =&- \int \big[(1 - \beta) \pa(z') + \beta \pb(z') \big] \frac{1}{\opt(z')} \delta(z - z') dz' - \lambda \int \delta(z - z') dz' \\
    =&- \big[(1 - \beta)\pa(z) + \beta \pb(z)\big] \frac{1}{\opt(z)} - \lambda = 0
\end{align}
Therefore
\begin{align}
    \opt(z) \propto \big[(1 - \beta)\pa(z) + \beta \pb(z)\big],
\end{align}
which corresponds to our $q$-path at $q=0$, or $\alpha = -1$ in \citet{amari2007integration}.  Thus, while both \citet{amari2007integration} and \citet{grosse2013annealing} start with the same objective, they arrive at different optimum because they optimize over different spaces.

\section{$q$-Exponential Families and Escort Moment-Averaging Path}\label{sec:student}\label{sec:parametric} \label{app:parametric}
In this section, we provide examples of parametric $q$-exponential family distributions and additional analysis for the special case of annealing between endpoints within the same parametric family.   After reviewing the $q$-Gaussian and Student-$t$ distributions as standard examples of the $q$-exponential family, we present the \textit{escort}-moments path, which is analogous to \citet{grosse2013annealing} and relies on the dual parameters of the $q$-family.    We experimentally evaluate these paths in toy examples in Fig. \ref{fig:toy_exp}, but note that the applicability of the escort-moments path is limited in practice.

\subsection{Examples of Parametric $q$-Exponential Family Distributions}
\paragraph{$q$-Gaussian and Student-$t$}
The $q$-Gaussian distribution appears throughout nonextensive thermodynamics \citep{naudts2009q, naudts2011generalised, tsallis2009introduction}, and corresponds to simply taking the $\exp_q$ of the familiar first and second moment sufficient statistics.   In what follows, we ignore the case of $q<1$ since the $q$-Gaussian has restricted support based on the value of $q$.  For $q > 1$, the $q$-Gaussian matches the Student-$t$ distribution, whose degrees of freedom parameter $\nu$ specifies the order of the $q$-exponential and introduces heavy tailed behavior.

The Student-$t$ distribution appears in hypothesis testing with finite samples, under the assumption that the sample mean follows a Gaussian distribution.   In particular, the degrees of freedom parameter $\nu = n-1$ can be shown to correspond to an order of the $\qqq$-exponential family with $\nu = (3-\qqq) / (\qqq-1)$ (in 1-d), so that the choice of $\qqq$ is linked to the amount of data observed.

We can first write the multivariate Student-$t$ density, specified by a mean vector $\mu$, covariance $\bSig$, and degrees of freedom parameter $\nu$, in $d$ dimensions, as
\begin{align}
    t_{\nu}(\vx | \bmu, \bSig) = \frac{1}{Z(\nu, \bSig)} \big[ 1 + \frac{1}{\nu} (x - \bmu)^T \bSig^{-1} (x-\bmu) \big]^{-\big(\frac{\nu + d }{2}\big)} \label{eq:student}
\end{align}
where $Z(\nu, \bSig) = \Gamma(\frac{\nu+d}{2})/\Gamma(\frac{\nu}{2}) \cdot |\bSig|^{-1/2} \nu^{-\frac{d}{2}} \pi^{-\frac{d}{2}}$.  Note that  $\nu > 0$, so that we only have positive values raised to the $-(\nu+d)/2$ power, and the density is defined on the real line.

The power function in \eqref{eq:student} is already reminiscent of the $\qqq$-exponential, while we have first and second moment sufficient statistics as in the Gaussian case.  We can solve for the exponent, or order parameter $q$, that corresponds to $-(\nu+d)/2$ using $-\big(\frac{\nu + d }{2}\big) = \frac{1}{1-\qqq}$.  This results in the relations
\begin{align}
 \nu = \frac{d - d \qqq +2}{\qqq - 1} \qquad \text{or} \qquad \qqq = \frac{\nu+d+2}{\nu+d}
\end{align}
We can also rewrite the $\nu^{-1} \, (x - \bmu)^T \bSig^{-1} (x-\bmu) $ using natural parameters corresponding to $\{x, x^2\}$ sufficient statistics as in the Gaussian case (see, e.g.
Matsuzoe and Wada (2015) Example 4).

Note that the Student-$t$ distribution has heavier tails than a standard Gaussian, and reduces to a multivariate Gaussian as $\qqq \rightarrow 1$ and $\exp_{\qqq}(u) \rightarrow \exp(u)$.  This corresponds to observing $n\rightarrow \infty$ samples, so that the sample mean and variance approach the ground truth \citep{murphy2007conjugate}. 

\paragraph{Pareto Distribution}
The $q$-exponential family can also be used for modeling the \textit{tail} behavior of a distribution \citep{bercher2008new, vehtari2015pareto}, or, in other words, the probability of $p(x)$ restricted to $X > x_{\text{min}} $ and normalized.

For example, the generalized Pareto distribution is defined via the tail function
\begin{align}
    P(X > x) = \begin{cases}  \big(1 + \xi \frac{x-x_{\text{min}} }{\sigma} \big)^{-\frac{1}{\xi}} \quad \xi \neq 0 \\  \exp\{- \frac{x-x_{\text{min}} }{\sigma}\} \qquad \xi =0
    \end{cases}
\end{align}
When $\xi \geq 0$,  the domain is restricted to $x \geq x_{\text{min}}$, whereas when $\xi < 0$, the support is between $x_{\text{min}} \leq x \leq x_{\text{min}}-\frac{\sigma}{\xi}$. Writing the CDF as $1- P(X>x)$ and differentiating leads to
\begin{align}
    p(x) = \frac{1}{\sigma}\big[1 + \xi \cdot \frac{x-x_{\text{min}}}{\sigma}  \big]^{-\frac{1}{\xi}-1}
\end{align}
Solving $-\frac{1}{\xi}-1 = \frac{1}{1-q}$ in the exponent, we obtain $q = \frac{2\xi + 1}{\xi+1}$ or $\xi = \frac{q-1}{q-2}$ .


\subsection{$q$-Paths between Endpoints in a Parametric Family}\label{app:same_family}
If the two endpoints $\pi_0, \tpi_1$ are within a $q$-exponential family, we can show that each intermediate distribution along the $q$-path of the same order is also within this $q$-family.  However, we cannot make such statements for general endpoint distributions, members of different $q$-exponential families, or $q$-paths which do not match the index of the endpoint $q$-parametric families.

\paragraph{Exponential Family Case}
We assume potentially vector valued parameters $\theta = \{ \theta\}_{i=1}^N$ with multiple sufficient statistics $\phi(\vz) = \{ \phi_i(\vz) \}_{i=1}^N$, with $\theta \cdot \phi(\vz) = \sum_{i=1}^N \theta_i \phi_i(\vz)$.
For a common base measure $\base(\vz)$, let $\pi_0(\vz) = \base(\vz) \, \exp\{ \theta_0 \cdot \phi(\vz) \}$ and $\tpi_1(\vz) = \base(\vz) \, \exp \{ \theta_1 \cdot \phi(\vz) \}$.   Taking the geometric mixture,
\begin{align}
    \tpi_\beta(\vz) &= \exp \big\{ (1-\beta) \, \log \pi_0(\vz) + \beta \, \log \tpi_1(\vz) \big\} \\
    &= \exp \big \{ \log \base(\vz) + (1-\beta) \, \theta_0 \cdot \phi(\vz) + \beta  \, \theta_1 \phi(\vz)  \big \} \\
    &= \base(\vz) \exp  \big \{ \big( (1-\beta) \, \theta_0 + \beta  \, \theta_1 \big) \cdot \phi(\vz)  \big \}
\end{align}
which, after normalization, will be a member of the exponential family with natural parameter $(1-\beta) \, \theta_0 + \beta  \, \theta_1$.

\paragraph{$q$-Exponential Family Case} For a common base measure $\base(\vz)$, let $\pi_0(\vz) = \base(\vz) \, \exp_q \{ \theta_0 \cdot \phi(\vz) \}$ and $\tpi_1(\vz) = \base(\vz) \, \exp_q \{ \theta_1 \cdot \phi(\vz) \}$.   The $q$-path intermediate density becomes
\begin{align}
\tilde{\pi}^{(q)}_\beta(\vz) &= \big[ (1-\beta) \, \pi_0(\vz)^{1-q} + \beta \, \tpi_1(\vz)^{1-q} \big]^{\frac{1}{1-q}} \\
&= \big[ (1-\beta) \, \base(\vz)^{1-q} \, \exp_q \{ \theta_0 \cdot \phi(\vz) \}^{1-q} + \beta \, \base(\vz)^{1-q} \,\exp_q \{ \theta_1 \cdot \phi(\vz) \} ^{1-q} \big]^{\frac{1}{1-q}} \\
&=  \bigg[ \base(\vz)^{1-q} \bigg ( (1-\beta) \,  \, [1 + (1-q)( \theta_0 \cdot \phi(\vz))]^{\frac{1}{1-q}1-q} + \beta  \, [1 + (1-q)( \theta_1 \cdot \phi(\vz))]^{\frac{1}{1-q} 1-q} \bigg) \bigg]^{\frac{1}{1-q}} \nonumber \\
&= \base(\vz) \bigg[ 1 + (1-q) \bigg( \big((1-\beta) \, \theta_0 + \beta  \, \theta_1 \big)  \cdot \phi(\vz) \bigg) \bigg]^{\frac{1}{1-q}} \\
&= \base(\vz) \exp_q \big\{  \big((1-\beta) \, \theta_0 + \beta  \, \theta_1 \big) \cdot \phi(\vz) \big\}
\end{align}
which has the form of an unnormalized $q$-exponential family density with parameter $(1-\beta) \, \theta_0 + \beta  \, \theta_1$.

\paragraph{Annealing between Student-$t$ Distributions}\label{app:student1d}
In \cref{fig:student_path}, we consider annealing between two 1-dimensional Student-$t$ distributions.  We set $q=2$, which corresponds to $\nu = 1$ with $\nu = (3-q) / (q-1)$, and use the same mean and variance as the Gaussian example in Fig. \ref{fig:alpha_path}, with $\pi_0(z) = t_{\nu=1}( -4, 3)$ and $\pi_1(z) = t_{\nu=1}( 4, 1)$.
For this special case of both endpoint distributions within a parametric family, we can ensure that the $q=2$ path stays within the $q$-exponential family of Student-$t$ distributions, just as the $q=1$ path stayed within the Gaussian family in Fig. \ref{fig:alpha_path}.

Comparing the $q=0.5$ and $q=0.9$ paths in the Gaussian case (Fig. \ref{fig:q_path}) with the $q=1.0$ and $q=1.5$ path for the Student-$t$ family with $q=2$, we observe that mixing behavior appears to depend on the relation between the $q$-path parameter and the order of the $q$-exponential family of the endpoints.  For our experiments in the main text, we did not find benefit to increasing $q>1$.  However, the toy example above indicates that $q>1$ may be useful in some settings, for example involving heavier tailed distributions.

As $q \rightarrow \infty$, the power mean \eqref{eq:abstract_mean} approaches the $\min$ operation as $1-q \rightarrow -\infty$.    In the Gaussian case in \cref{fig:q_path}, we see that, even at $q=2$, intermediate densities for all $\beta$ appear to concentrate in regions of low density under both $\pi_0$ and $\pi_T$.   However, for the heavier-tailed Student-$t$ distributions, we must raise the $q$-path parameter significantly to observe similar behavior.

\begin{figure}[t]
    \centering
    \includegraphics[trim={0 0 0 0 },clip,width=0.99\textwidth]{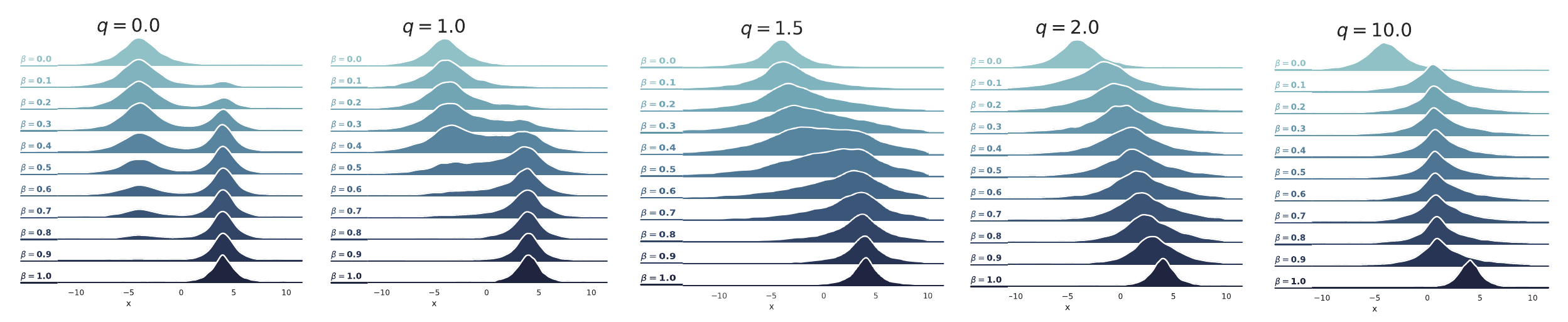}
    \caption{Intermediate densities between Student-$t$ distributions, $t_{\nu = 1}(-4, 3)$ and $t_{\nu = 1}(4,1)$ for various $q$-paths and 10 equally spaced $\beta$,
    Note that $\nu=1$ corresponds to $q=2$, so that the $q=2$ path stays within the $q$-exponential family.}
    \label{fig:alpha_path2} 
    \label{fig:student_path} 
    \vspace*{-.15cm}
\end{figure}

\subsection{Moment-Averaged Path as a Generalized Mean}\label{app:moments_as_generalized_mean}
While our $q$-paths can take arbitrary unnormalized density functions ${\textbf{u} = \big( \tpi_0(\vz), \tpi_1(\vz) \big)}$ as input arguments for the generalized mean,  we can reinterpret the moment-averaging path as a generalized mean over the natural parameters ${\textbf{u} = \big( \theta_0, \theta_1  \big)}$.   We contrast the difficulty of inverting the function $h(\theta)$ for the moments path (which involves the Legendre transform), against the simple form of the geometric or $q$-paths as arithmetic means in the parameter space $\theta$ as in \cref{app:same_family}.
The moment-averaged path is defined using a convex combination of the dual parameter vectors \citep{grosse2013annealing}, for the restricted case where $\pi_0(\vz)$ and $\pi_1(\vz)$ are members of the same exponential family, with parameters $\theta_0$ and $\theta_1$ 
\begin{align}
\eta(\theta_{\beta}) = (1-\beta) \, \eta(\theta_0) + \beta \, \eta(\theta_1) \, . \label{eq:moments_path}
\end{align}
To solve for the corresponding natural parameters, we calculate the Legendre transform, or a function inversion $\eta^{-1}$.
\begin{align}
    \theta_{\beta} = \eta^{-1}\big((1-\beta) \, \eta(\theta_0) + \beta \, \eta(\theta_1)\big) \label{eq:moments_path_theta} \,.
\end{align}
 Comparing to the form of  \cref{eq:abstract_mean}, we can interpret the moment-averaging path as a generalized mean, with the natural parameters $\bf{u} = (\theta_0, \theta_1)$ as inputs and the sufficient statistic function as the transformation $h(\theta) = \eta(\theta)$, although calculating the inverse is difficult in practice.

 This observation highlights the convenience of working with generalized means in unnormalized density function space as in $q$-paths.   When constructing paths from generalized means in parameter space $\theta$, one may have to calculate normalization constants or consider the entire domain of the density function.  By contrast, the expression for $q$-paths in \cref{eq:qpath_mix_form} only involves inverting a scalar function at each point in the input sample space $\vz$.


\subsection{Escort Moment-Averaged Path}
While exponential families are ubiquitous throughout machine learning, whether via common parametric distributions such as Gaussians or energy-based models such as (Restricted) Boltzmann Machines, models involving the $q$-exponential function have received comparatively little attention in machine learning.
Nevertheless, we derive an analogue of the moment-averaged path for endpoint distributions within the same $q$-exponential family, with several parametric examples in App. \ref{app:parametric}.   We begin by recalling the definition,
\begin{align}
    \pi_{\theta,q}(\vz) &= g(\vz) \, \exp_q \big\{ \theta \cdot \suffexp_q(\vz) - \psi_q(\theta ) \big\}. \label{eq:qexp_fam}
\end{align}
where $g(z)$ indicates a base distribution and $\psi_q(\theta)$ denotes the $q$-free energy, which is convex as a function of the parameter $\theta$ \citep{amari2011q}.


As in the case of the exponential family, differentiating the $q$-free energy yields a dual parameterization of the $q$-exponential family \citep{amari2011q}.  However, the standard expectation is now replaced with the \textit{escort} expectation \citep{naudts2011generalised}
\begin{align}
    \eta_q(\theta) = \nabla_\theta \psi_q(\theta) &= \int \frac{\tpi_\theta^{(q)}(\vz)^{q}}{\int \tpi_\theta^{(q)}(\vz)^{q}} \cdot \phi(\vz) d\vz \\
    &:= \mathbb{E}_{\Pi_q(\theta)}[ \phi(\vz)] 
\end{align}
where $\Pi_q(\theta) \propto \tpi_{\theta,q}(\vz)^{q}$ is the escort distribution for a given for $\tpi_{\theta,q}$ in a parametric $q$-exponential family.   This reduces to the standard expectation for $q=1$ as in \cref{eq:dpsi_dtheta}.

We propose the escort moment-averaging path for endpoints within a $q$-exponential family, using linear mixing in the dual parameters.  Letting the function $\eta_{\Pi}(\theta)$ output the escort expected sufficient statistics for a $q$-exponential family distribution with parameter $\theta$,
\begin{align}
    \eta_{\Pi_q}(\theta_{\beta}) = (1-\beta) \, \eta_{\Pi_q}(\theta_0) + \beta \, \eta_{\Pi_q}(\theta_1)
\end{align}

To provide a concrete example of the escort moment-averaging path in Fig. \ref{fig:student_path}, we consider the Student-$t$ distribution, which uses the same first- and second-order sufficient statistics as a Gaussian distribution and a degrees of freedom parameter $\nu$ that specifies the order of the $q$-exponential function for $q \geq 1$.  This parameter induces heavier tails than a standard Gaussian, which appears as a special case as $q \rightarrow 1$ and $\exp_{q}(u) \rightarrow \exp(u)$.

In Fig. \ref{fig:student_path}, we observe that the escort moments path spreads probability mass more widely than the $q$-path, which matches the observations of \citet{grosse2013annealing} in comparing the moment-averaging path to the geometric path for exponential family endpoints.
Note that the $q$-path remains within the $q$-exponential family as shown in \cref{app:same_family}.

We proceed to derive a closed form expression for the parameters of intermediate distributions along the escort moment-averaged path between Student-$t$ endpoints.

\begin{figure*}[t]
    \centering
    \includegraphics[trim={0 0 0 0 },clip,width=0.99\textwidth]{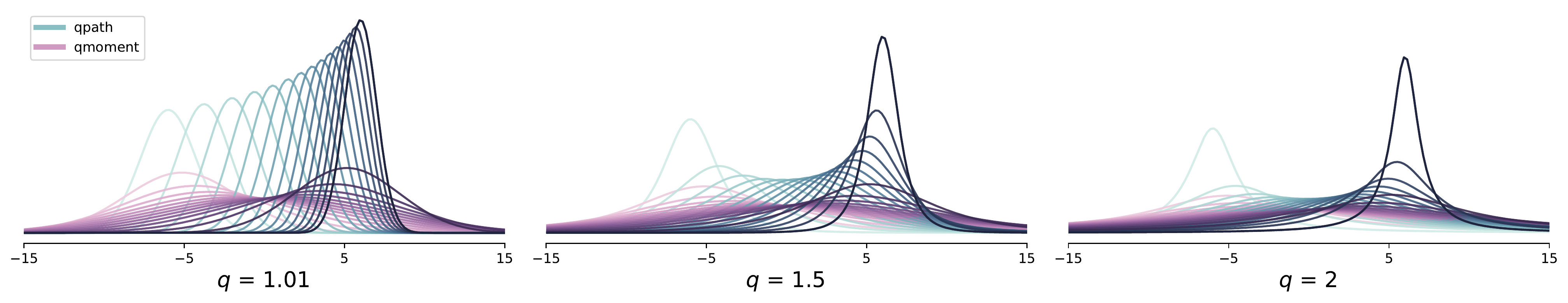}
    \caption{We visualize the escort-moments path for Student-$t$ endpoints with $t_{\nu}(-4, 3)$ and $t_{\nu}(4,1)$ for various $\nu = (3-q)/(1-q)$.  We compare the corresponding $q$-path, whose intermediate densities remain within the $q$-exponential family, to the escort-moments path (\cref{eq:escort}).  Note, $q=1.01$ closely resembles the moment-averaged path of \citet{grosse2013annealing}.}
    \label{fig:alpha_path}
    \label{fig:student_path} 
    \vspace*{-.15cm}
\end{figure*}

\subsection{Escort Moment-Averaged Path with Student-$t$ Endpoints}\label{app:escort_student}

For the case of the Student-$t$ distribution with degrees of freedom parameter $\nu$ , the escort distribution is \textit{also} a Student-$t$ distribution, but with $\nu^\prime= \nu + 2$ and a rescaling of the covariance matrix $\frac{1}{Z_{\Pi}(\Sigma)} t_{\nu}(\vz ;  \mu, \Sigma)^{q} = t_{\nu+2}(\vz ; \mu, \frac{\nu}{\nu+2} \Sigma)$ 
(Tanaka 2010, Matsuzoe 2017).

Finding the escort moment-averaged path thus becomes a moment matching problem over Student-$t$ distributions with a different $\nu$.  We seek to find $\pi_\beta(\vz) = t_{\nu}(\vz ;\mu_{\beta}, \Sigma_{\beta})$ such that the expected sufficient statistics,  under the escort distribution $\Pi_\beta(\vz) = t_{\nu+2}(\vz ; \mu_\beta, \frac{\nu}{\nu+2} \Sigma_\beta)$, are equal to
\begin{align}
   \mathbb{E}_{\Pi_\beta}\left[ \vz \right] &= (1-\beta) \, \mathbb{E}_{\Pi_0}\left[ \vz \right] + \beta \, \mathbb{E}_{\Pi_1}\left[ \vz \right] \\
    \mathbb{E}_{\Pi_\beta}\left[ \vz \vz^T \right] &= (1-\beta) \, \mathbb{E}_{\Pi_0}\left[ \vz \vz^T \right] + \beta \, \mathbb{E}_{\Pi_1}\left[ \vz \vz^T \right]
\end{align}
where optimization is over the parameters of the distribution $t_{\nu}(\vz ; \mu_{\beta}, \Sigma_{\beta})$.   Note that $\mathbb{E}_{\Pi_\beta}\left[ \vz \right] = \mu_{\beta}$ since the mean is unchanged for the escort distribution, whereas $\mathbb{E}_{\Pi_\beta}\left[ \vz \vz^T \right] = \Sigma_{\Pi_{\beta}} + \mu_{\Pi_{\beta}} \mu_{\Pi_{\beta}}^T = \frac{\nu}{\nu+2} \Sigma_{\beta} + \mu_{\beta} \mu_{\beta}^T$.




Following similar derivations as in \citet{grosse2013annealing} Sec. 4 using the escort expressions, we have
\begin{align}
\mu_{\beta}= \mu_{\Pi_{\beta}}  &= (1-\beta) \mu_0 + \beta \mu_1 \label{eq:escort}\\
\Sigma_{\beta} =  \frac{\nu+2}{\nu} \Sigma_{\Pi_{\beta}}  &= (1-\beta) \Sigma_0 + \beta \Sigma_1  + \frac{\nu+2}{\nu} \beta (1-\beta) (\mu_1 - \mu_0)(\mu_1-\mu_0)^T \nonumber
\end{align}
which implies that the escort moment-averaged distribution has the form $t_{\nu}(\vz; \mu_{\beta}, \Sigma_{\beta})$, with the same degrees of freedom $\nu$ as in the original $q$-exponential family.

\section{Additional Experiments for Parametric Endpoint Distributions  }

In these experiments, we consider using \gls{AIS} to estimate the partition function ratio for well-separated 1-d Gaussian ($q=1$) and Student-$t$ ($q>1$) endpoint distributions.    Our goal is to compare the performance of the moment-averaging or escort-moment averaging paths, which are limited to the case of parametric endpoints distributions, with the more general $q$-paths.

\paragraph{Gaussian}
To compare $q$-paths against the moment-averaging path \citep{grosse2013annealing}, we anneal between $\pi_0=\mathcal{N}(-4,3)$ and $\pi_1 = \mathcal{N}(4,1)$.   Similarly, we anneal between $\pi_0=t_{\nu=1}(-4, 3)$ and $\pi_1 = t_{\nu=1}(4, 1)$, where $\nu = 1$ corresponds to $q=2$, to compare against the escort moment-averaged path in \cref{sec:student}.   For all experiments, we use use parallel runs of \gls{HMC} \citep{neal2011mcmc} to obtain 2.5k independent samples from $\tilde{\pi}_{\beta, q}(\vz)$ using $K$ linearly spaced $\beta_t$ between $\beta_0=0$ and $\beta_K=1$. We perform a grid search over 20 log-spaced $\delta \in [10^{-5}, 10^{-1}]$ and report the best $q = 1 - \delta$.

\begin{figure*}[h]
    \centering
    \subfigure[$\mathcal{N}(-4,3) \rightarrow \mathcal{N}(4,1)$]{\includegraphics[trim={0 0cm 0 0cm},clip,width=0.42\textwidth]{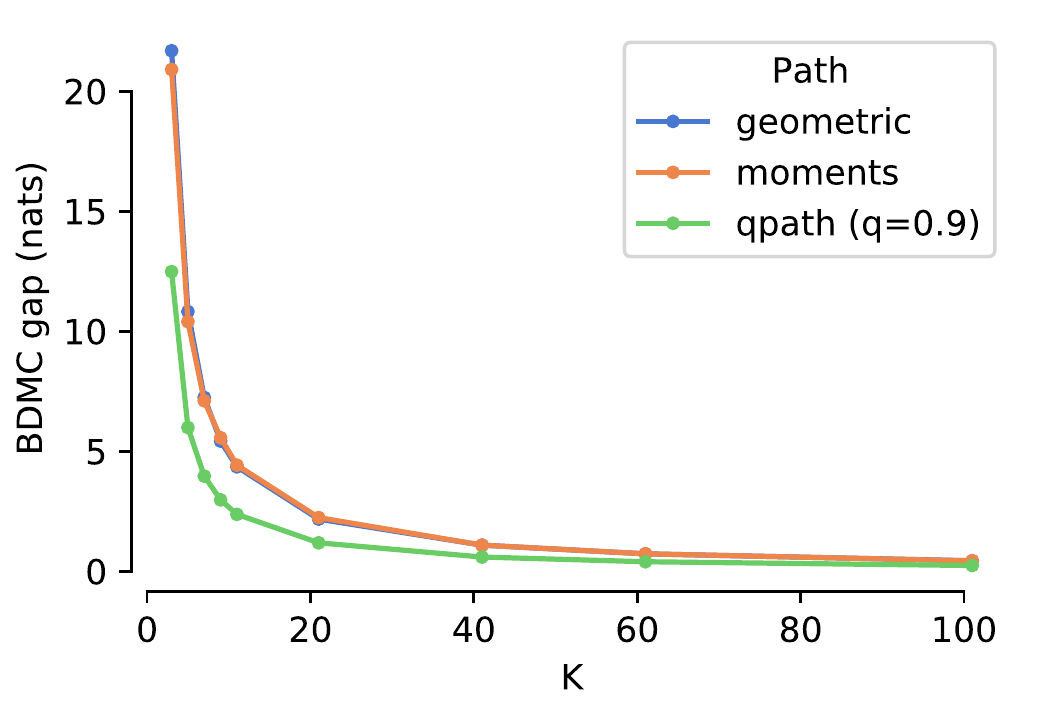}}
    \subfigure[$t_{v=1}(-4, 3) \rightarrow t_{v=1}(4,1)$]{\includegraphics[trim={0 0cm 0 0cm},clip,width=0.42\textwidth]{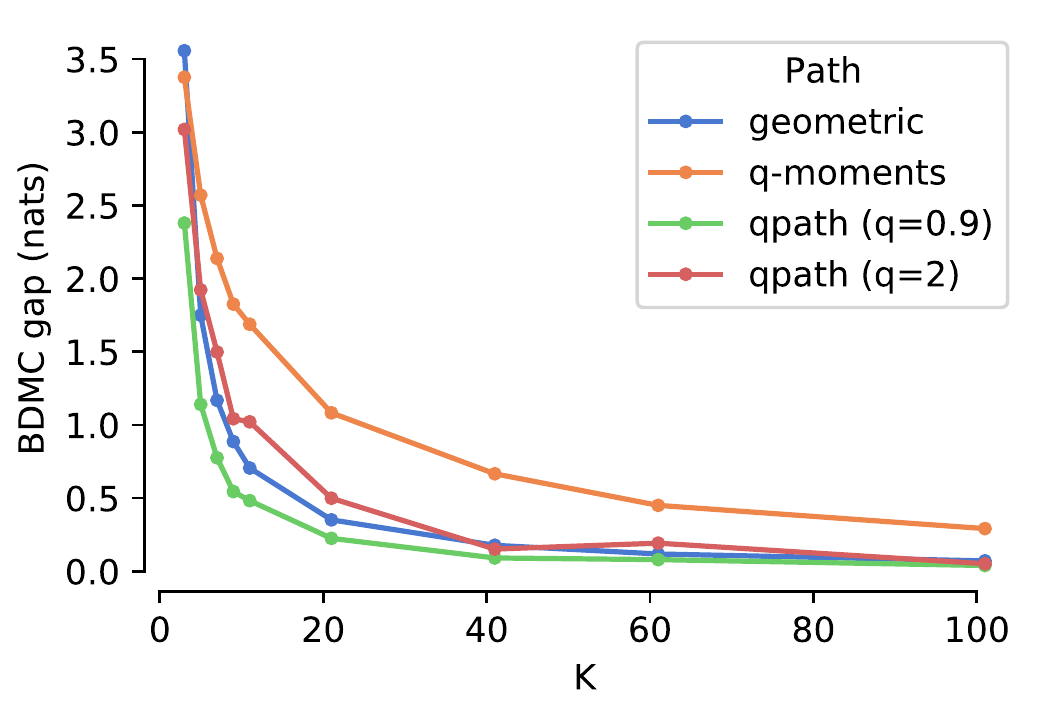}}
    \vspace*{-.25cm}
    \caption{\acs{BDMC} gaps for various paths on toy models. $q$-Paths out perform both the moments and the escort-moments path, both of which make use of parametric endpoint assumptions. Best $q$ out of 20 shown.}
    \label{fig:toy_exp}
\end{figure*}

Results are shown in \cref{fig:toy_exp}, where we observe $q$-paths outperform the geometric path in both cases, as well as the moment and $q$-moments paths which have closed-form expressions and exact samples.   In App. \ref{app:student1d}, we provide additional analysis for annealing between two Student-$t$ distributions.

\paragraph{Student-$t$}
Since the Student-$t$ family generalizes the Gaussian distribution to $q \neq 1$, we can run a similar experiment annealing between two Student-$t$ distributions.   We set $q=2$, which corresponds to $\nu = 1$ with $\nu = (3-q) / (q-1)$, and use the same mean and variance as the Gaussian example in Fig. \ref{fig:alpha_path} or Student-$t$ example in Fig. \ref{fig:student_path} with $\pi_0(z) = t_{\nu=1}( -4, 3)$ and $\pi_1(z) = t_{\nu=1}( 4, 1)$.

In \cref{fig:toy_exp}, we compare the escort-moment averaging path with $q=2$ to the geometric path and various $q$-paths.   As shown in \cref{app:same_family}, the $q$-path with $q=2$ stays within the $q$-exponential family.   The escort-moment averaging path does not outperform $q$-paths, which may be surprising since it appears to achieve interesting mass covering behavior in Fig. \ref{fig:student_path}.   As in the Gaussian case, we see that $q$-paths with $q \neq 2$ can achieve improvements even when the endpoints Student-$t$ distributions use $q=2$.

\section{Sum and Product Identities for $q$-Exponentials}\label{app:q_sum_product}
In this section, we prove two lemmas which are useful for manipulation expressions involving $q$-exponentials, for example in moving between \cref{eq:normalization1} and \cref{eq:normalization2} in either direction.
\begin{lemma}
    Sum identity
    \begin{align}
        \exp_q\left(\sum_{n=1}^N x_n\right) = \prod_{n=1}^{N} \exp_q \left(\frac{x_n}{1 + (1 - q)\sum_{i=1}^{n-1}x_i} \right)\label{eq:q_exp_sum}
    \end{align}
    \label{lemma:q_exp_sum}
\end{lemma}
\begin{lemma}
    Product identity
    \begin{align}
        \prod_{n=1}^N \exp_q(x_n) = \exp_q\left(\sum_{n=1}^{N}x_n \cdot \prod_{i=1}^{n-1} \left(1 + (1 - q)x_i\right)\right)\label{eq:q_exp_prod}
    \end{align}
    \label{lemma:q_exp_prod}
\end{lemma}

\subsection{Proof of Lemma 1}
\begin{proof}
    We prove by induction. The base case ($N=1$) is satisfied using the convention $\sum_{i=a}^bx_i = 0$ if $b < a$ so that the denominator on the \textsc{rhs} of \cref{eq:q_exp_sum} is $1$. Assuming \cref{eq:q_exp_sum} holds for $N$,
    \begin{align}
        \exp_q\left(\sum_{n=1}^{N+1} x_n\right) &= \left[ 1 + (1-q) \sum_{n=1}^{N+1} x_n \right]_{+}^{1/(1-q)} \\
                                                &= \left[ 1 + (1-q) \left(\sum_{n=1}^{N} x_n\right) + (1-q)x_{N+1} \right]_{+}^{1/(1-q)} \\
                                                &= \left[\left( 1 + (1-q) \sum_{n=1}^{N} x_n \right) \left(1 + (1-q)\frac{x_{N+1}}{1 + (1-q) \sum_{n=1}^{N} x_n}\right) \right]_{+}^{1/(1-q)} \\
                                                &= \exp_q\left(\sum_{n=1}^N x_n\right) \exp_q \left(\frac{x_{N+1}}{1 + (1-q) \sum_{n=1}^{N} x_n} \right)\\
                                                &= \prod_{n=1}^{N+1} \exp_q \left(\frac{x_n}{1 + (1-q)\sum_{i=1}^{n-1}x_i} \right) \text{(using the inductive hypothesis)}
    \end{align}
\end{proof}
\subsection{Proof of Lemma 2} 
\begin{proof}
    We prove by induction. The base case ($N=1$) is satisfied using the convention $\prod_{i=a}^bx_i = 1$ if $b < a$. Assuming \cref{eq:q_exp_prod} holds for $N$, we will show the $N+1$ case. To simplify notation we define $y_N:=\sum_{n=1}^{N}x_n \cdot \prod_{i=1}^{n-1} \left(1 + (1 - q)x_i\right)$. Then,
\begin{align}
    \prod_{n=1}^{N+1} \exp_q(x_n) &= \exp_q(x_{1})\left(\prod_{n=2}^{N+1}\exp_q(x_n)\right)\\
    &= \exp_q(x_{0})\left(\prod_{n=1}^{N}\exp_q(x_n)\right) & \hspace*{-.5cm} \text{(reindex $n \to n - 1)$} \nonumber \\
    &=\exp_q(x_{0})\exp_q(y_N) & \hspace*{-.5cm} \text{(inductive hypothesis)} \nonumber \\
    &= \bigg[\left(1 + (1-q) \cdot x_{0}\right)\left(1 + (1-q) \cdot y_N\right) \bigg]_{+}^{1/(1-q)}\\
    &= \bigg[1 + (1-q) \cdot x_{0} + \big(1 + (1-q) \cdot x_{0} \big)(1-q) \cdot y_N \bigg]_{+}^{1/(1-q)}\\
    &= \bigg[1 + (1-q) \bigg(x_{0} + \big(1 + (1-q) \cdot x_{0} \big)y_N\bigg) \bigg]_{+}^{1/(1-q)}\\
    &= \exp_q \left(x_{0} + \big(1 + (1-q) \cdot x_{0} \big)y_N\right)
\end{align}
Next we use the definition of $y_N$ and rearrange
\begin{align}
    &= \exp_q \left(x_{0} + \big(1 + (1-q) \cdot x_{0} \big)\left(x_1 + x_2(1 + (1-q) \cdot x_1) + ... + x_N \cdot \prod_{i=1}^{N-1}(1 + (1-q) \cdot x_i)\right)\right) \nonumber\\
    &= \exp_q\left(\sum_{n=0}^{N}x_n \cdot \prod_{i=1}^{n-1} \left(1 + (1-q) x_i\right)\right).
\end{align}
Then reindexing $n \to n + 1$ establishes
\begin{align}
    \prod_{n=1}^{N+1} \exp_q(x_n) = \exp_q\left(\sum_{n=1}^{N+1}x_n \cdot \prod_{i=1}^{n-1} \left(1 + (1-q)x_i\right)\right).
\end{align}
\end{proof}

\section{Experimental details and results}\label{sec:experiment_details}

\begin{algorithm}[h]
    \caption{ESS Heuristic for Q-paths}
\begin{spacing}{1.2}
 \begin{algorithmic}[1] \label{alg:ess_heuristic}
    \STATE {\bfseries Input:} Set of log weights $\{\log w_i\}_{i=1}^S$, random restarts $M$, sample variance $\sigma$
    \STATE {\bfseries Output:} $q, \beta$ which minimizes ESS criterion from \citet{chopin2020introduction}.
    \STATE Initialize $\delta_{0} = \max_i|\log w_i|$ and $\mathcal{L}_{\text{best}} = \infty $\\
    \FOR{$j$ from $1$ to $M$}
        \STATE Initialize $\beta_{0} = 1$, $q_0 = 1 - \rho^{-1}$ with $\rho \sim \mathcal{N}(\rho_0, \sigma)$
        \STATE Solve $\beta^*, q^* = \argmin_{\beta, q} \mathcal{L}(\beta_0, q_0)$ with $\mathcal{L}$ defined in \cref{eq:ess_loss} using coordinate descent.
        \IF{$\mathcal{L}(\beta^*, q^*) < \mathcal{L}_{\text{best}}$}
            \STATE Set $q_{\text{best}} \leftarrow q^*$, $\beta_{\text{best}} \leftarrow \beta^*$, $\mathcal{L}_{\text{best}} \leftarrow \mathcal{L}(\beta^*, q^*)$
        \ENDIF
    \ENDFOR{}
    \STATE {\bfseries return} $q_{\text{best}}, \beta_{\text{best}}$
 \end{algorithmic}
\end{spacing}
\end{algorithm}

\subsection{Sequential Monte Carlo}\label{app:smc_exp_details}

We follow the experimental setup from Ch. 17.3 of \citet{chopin2020introduction} using the preprocessed Pima Indians diabetes ($N=768, D=9$) and Sonar datasets ($N=208, D=61$) available at \url{https://particles-sequential-monte-carlo-in-python.readthedocs.io/en/latest/datasets.html}. The model is specified as:
\begin{align}
    p(w_j) &= \mathcal{N}(0, 5^2) \quad \quad
    p(y_i|x_i, w) = \text{Bern}(p_i = \text{sigmoid}(x_i^Tw)) \\
    p(\theta) &= \prod_{j=1}^Dp(w_j) \quad \quad
    p(\mathcal{D}, \theta) = p(\theta)\prod_{i=1}^Np(y_i|x_i, w).
\end{align}
In \cref{alg:ess_heuristic} we use $M=100$ restarts and compute $\rho$ in $\log_{10}$ space with a sample variance $\sigma = 0.1$ (i.e $q = 1 - 10^{-\rho}$ for $\rho \sim \mathcal{N}(\log_{10}{(\rho_0)}, 0.1)$). For coordinate descent we use the modified Powell algorithm available from the scipy python library.

\subsection{Evaluating generative models using AIS}\label{app:ais_exp_details}

\begin{table}[!ht]
\centering
\caption{Settings for training and evaluating a \gls{VAE} generative model trained with \gls{TVO} on the Omniglot dataset.}
\begin{tabular}{c|c}
\textbf{Configuration} & \textbf{Value}          \\ \hline
training examples      & 24,345                  \\
simulated examples     & 2,500                   \\
real test examples     & 8,070                  \\ \midrule
epochs                 & 5000                    \\
number of importance samples & 50                       \\
number of TVO partitions & 100 \\
TVO partition schedule & log uniform ($\beta_1=0.025$)\\
decoder                & {[}50, 200, 200, 784{]} \\
encoder                & {[}784, 200, 200, 50{]} \\
batch size             & 100                     \\
activation function    & tanh
\end{tabular}
\end{table}

\begin{figure}[t]
    \centering
    \includegraphics[width=0.8\textwidth]{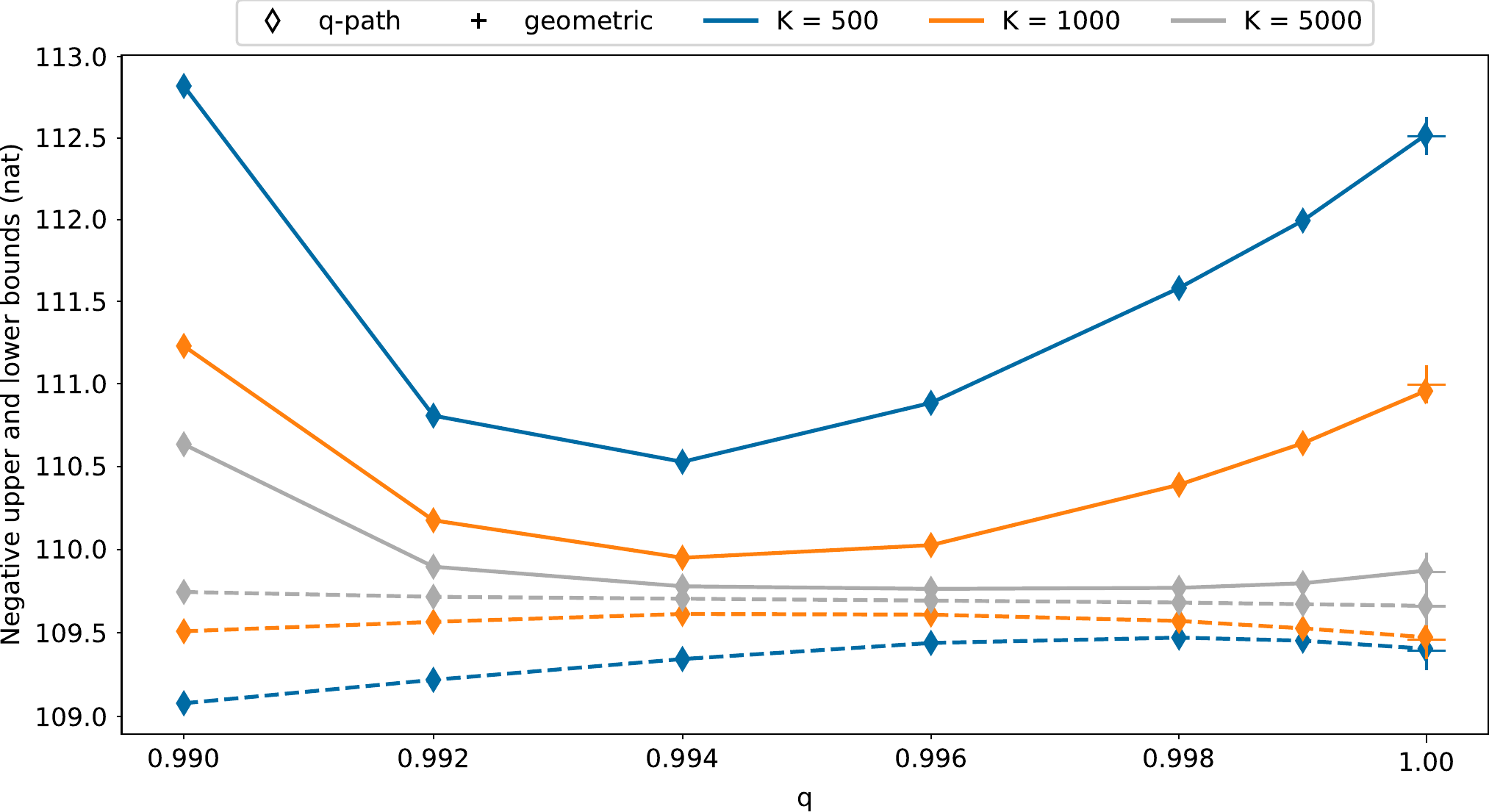}
    \caption{Stochastic lower and upper bounds produced by forward and reverse Hamiltonian AIS runs, for various numbers of annealing distributions ($K$) and $q$-values. Best viewed in colour. \label{fig:bdmc_omniglot_bounds}}
\end{figure}


\end{document}